\pdfoutput=1
\documentclass[journal]{IEEEtran}

\usepackage{algorithm,algorithmic}
\usepackage{amsfonts}
\usepackage[pdftex]{graphicx}
\usepackage{tikz}
\usepackage{multirow}

\usepackage{amssymb}
\usepackage{enumitem}
\usepackage{layouts}
\usepackage{siunitx}
\usepackage{textcomp}
\usepackage[flushleft]{threeparttable}
\usepackage{booktabs}
\usepackage[tbtags]{amsmath}
\usepackage[pdfencoding=auto, psdextra]{hyperref}
\usepackage[nameinlink,noabbrev,capitalise]{cleveref}
\crefformat{equation}{(#2#1#3)}
\usepackage{bookmark}
\usepackage{tikz}
\usepackage{tabularx}
\newcolumntype{Y}{>{\centering\arraybackslash}X}
\usepackage{bm}

\newcommand\widebar[1]{\mathop{\overline{#1}}}

\usepackage{xcolor,colortbl}

\definecolor{Gray}{gray}{0.85}
\definecolor{LightCyan}{rgb}{0.88,1,1}

\newcolumntype{a}{>{\columncolor{Gray}}c}
\newcolumntype{b}{>{\columncolor{white}}c}

\definecolor{cobalt}{rgb}{0.0, 0.28, 0.67}
\definecolor{coolblack}{rgb}{0.0, 0.18, 0.39}

\usepackage{amsmath}
\ifCLASSINFOpdf

\else

\fi

\ifCLASSOPTIONcompsoc
 \usepackage[caption=false,font=normalsize,labelfont=sf,textfont=sf]{subfig}
\else
 \usepackage[caption=false,font=footnotesize]{subfig}
\fi

\begin{document}
\title{Multiple-Identity Image Attacks Against\\Face-based Identity Verification}

\author{Jerone~T.~A.~Andrews,
        Thomas~Tanay,
        and~Lewis~D.~Griffin
\thanks{J. T. A. Andrews (e-mail: jerone.andrews@cs.ucl.ac.uk) is with the Department of Computer Science, University College London, London, UK; and Telefónica Research, Barcelona, ES.}
\thanks{T. Tanay is with the Huawei Noah’s Ark Lab, London, UK (work done while at University College London).}
\thanks{L. D. Griffin is with the Department of Computer Science, University College London, London, UK.}
\thanks{J. T. A. Andrews is supported by the Royal Academy of Engineering (RAEng) and the Office of the Chief Science Adviser for National Security under the UK Intelligence Community Postdoctoral Fellowship Programme; and acknowledges further research funding from the European Union’s Horizon 2020 research and innovation programme under the Marie Skłodowska-Curie ENCASE project, Grant Agreement No. 691025.}}


\maketitle

\begin{abstract}
Facial verification systems are vulnerable to poisoning attacks that make use of multiple-identity images (MIIs)---face images stored in a database that resemble multiple persons, such that novel images of any of the constituent persons are verified as matching the identity of the MII. Research on this mode of attack has focused on defence by detection, with no explanation as to why the vulnerability exists. New quantitative results are presented that support an explanation in terms of the geometry of the representations spaces used by the verification systems. In the spherical geometry of those spaces, the angular distance distributions of matching and non-matching pairs of face representations are only modestly separated, approximately centred at 90 and 40-60 degrees, respectively. This is sufficient for open-set verification on normal data but provides an opportunity for MII attacks. Our analysis considers ideal MII algorithms, demonstrating that, if realisable, they would deliver faces roughly 45 degrees from their constituent faces, thus classed as matching them. We study the performance of three methods for MII generation---gallery search, image space morphing, and representation space inversion---and show that the latter two realise the ideal well enough to produce effective attacks, while the former could succeed but only with an implausibly large gallery to search. Gallery search and inversion MIIs depend on having access to a facial comparator, for optimisation, but our results show that these attacks can still be effective when attacking disparate comparators, thus securing a deployed comparator is an insufficient defence.
\end{abstract}


\IEEEpeerreviewmaketitle

\section{Introduction}
\label{sec:intro}
\IEEEPARstart{B}{iometric} identifiers, such as face images, are commonly employed to verify the identity of individuals---capitalising on the distinctiveness of the identifier, its stability of appearance, and its fixed linkage with the identity. Recent advances in deep learning have enabled the development of algorithms that compare live and stored reference face images to automate this verification~\cite{schroff2015facenet, parkhi2015deep, liu2017sphereface,cao2018vggface2,wu2018light}. Increasingly such algorithms are being used for access control (e.g. international border crossing, banking access, and facility entry)~\cite{jain201650}.

An adversary attempting to defeat a face-based identity verification system has three options~\cite{ratha2001enhancing}: (i) avoid the system; (ii) disguise the live face; or (iii) poison the stored face-identity pairs. This paper is concerned with poisoning attacks.

Consider a database of stored face images. The simplest form of poisoning attack is to contrive an incorrect pairing: say the face image of \emph{A} paired with the identity of \emph{B}. This allows individual \emph{A} to pass as individual \emph{B}; which can serve either the aim of pretending to be \emph{B}, or avoiding notice as being \emph{A}. Achieving this incorrect pairing should be difficult though, as generally \emph{A} and \emph{B} will look different. Furthermore, great care is typically taken when establishing the pairings in the database---for example, by requiring trusted individuals to attest that the stored image is a good likeness of the person with whom it is to be paired.

A subtle way to bypass the safeguards against poisoning makes use of multiple-identity images (MIIs), which are face images that resemble more than one person. Assuming, for now, that MIIs are possible we explain how they would be used. Suppose \emph{A} wishes to access some facility without their identity being detected. \emph{A} works with an accomplice \emph{B} to prepare an MII resembling both \emph{A} and \emph{B}, and arranges for it to be stored in the database paired with the identity of \emph{B}. This storage could be achieved by creating a new entry in the database or by updating an old one for \emph{B}---in either case the safeguards against poisoning are circumvented since the MII resembles \emph{B}. Having poisoned the database, \emph{A} will now be able to access the facility undetected as their live face will match the stored MII and they will be verified as \emph{B}.

Verification systems are used in a variety scenarios with differing restrictions on the capture of reference and live images. The most restricted type are exemplified by passport photographs which can be captured only under constrained conditions of pose, illumination and expression. At intermediate level are systems that use face photographs captured by a webcam or mobile phone. The least restricted type are walk past systems, collecting face images without subject participation. In this paper, we examine the issue of MIIs, where the verification system is trained for use in the unconstrained setting.

The aims of this paper are: (i) to explain why MIIs are possible in principle; (ii) to show that MIIs can be realised well enough to be used practically; and (iii) to show that securing a face comparator does not prevent these attacks.

Our analyses will make use of the representation spaces that are used by automated methods of face verification. For the first aim, we will investigate the distribution of faces within a representation space. For the second, we will present one method that constructs MIIs directly in image space, and a second method that constructs MIIs first in representation space and then in image space. For the third, we will show that the representation spaces of different comparators are sufficiently similar to permit the transfer of MIIs.

In \Cref{sec:rel_work} we give an overview of related work. In \Cref{sec:prelim} we introduce the datasets and face comparators that we experiment with. In \Cref{sec:why_vul} we offer an explanation as to why face comparators are theoretically vulnerable to MIIs; and consider the behaviour and performance of a hypothetical \emph{ideal} method for MII generation. Subsequently, in \Cref{sec:im_space_attacks}, we outline three methods for either \emph{finding}, \emph{constructing}, or \emph{synthesising} MIIs---showing that real MIIs can be generated that are sufficiently close to ideal MIIs that they are effective attacks. In \Cref{sec:effectiveness} we evaluate the effectiveness of generated MIIs, as well as the transferability of MIIs to novel comparators. In \Cref{sec:discussion} we discuss the inherent vulnerability of face comparators to MIIs in light of our findings; and conclude in \Cref{sec:conclusion}.

\section{Background}
\label{sec:rel_work}
In this section, we provide an overview of input attack generation and input attack detection.

\subsection{Input Attack Generation}
Here we review two distinct paradigms for input attack generation: image space and representation space. Both were originally developed for innocuous tasks---in the creative industries, but have subsequently been re-purposed for MII generation.

\subsubsection{Image Space}
\label{subsubsec:imagespace}
Image morphing is the process of generating a sequence of photorealistic interpolations between two images~\cite{wolberg1998image}. Seminally, \cite{ferrara2014magic} demonstrated the vulnerability of two commercially available face verification systems to MII attacks generated using morphing. The morphs in that case were constructed by geometrically warping~\cite{wolberg1998image} and then colour interpolating two face images. Attacks of this type have been shown to be highly effective~\cite{scherhag2017vulnerability}, even deceiving human experts~\cite{robertson2018detecting}. 

The warping step is crucial, and reliant on precisely identified common landmarks. This approach to face morphing thus works best when both images are frontally aligned, but even then there are often cues to manipulation. Several works have attempted to improve the visual quality of facial morphs, in particular, by: manual replacement and movement of correspondence points~\cite{seibold2018reflection}; manual retouching \cite{ferrara2014magic}; splice morphing~\cite{makrushin2017automatic}; the restrictive selection of similar input images~\cite{vyas2015automatic}; and Poisson image editing~\cite{perez2003poisson}. Although successful in creating an image that matches two identities, at present image space morphing still tends to leave artefacts that cue the manipulation~\cite{debiasi2018prnu,seibold2018reflection}. 

\subsubsection{Representation Space}
A very different approach to image generation utilises algorithms that can synthesise a realistic image from a representation space encoding. Significantly, \cite{mahendran2015understanding,dosovitskiy2016inverting} found that when inverting deep representations, several layers preserve accurate image-specific information. For example, \cite{dosovitskiy2016generating} utilised a deconvolutional neural network, and adversarial training, to generate high-quality images given high-level representations. Similarly, ~\cite{shu2017neural} proposed to leverage approximate models of facial appearance for training a generative model under adversarial supervision. 

With attribute-conditioned image modification in mind (e.g. changing the age or gender of a face), invertible conditional generative adversarial networks \cite{perarnau2016invertible} have also been proposed. They task an encoder network with learning an inverse mapping from an image to both its latent and conditional attribute representation, thus permitting the modification of images by varying the conditional information. More recently, \cite{lample2017fader} proposed an encoder-decoder scheme that forces the latent representation to be invariant to the attributes. However, \cite{he2017arbitrary} argue that decoupling the attribute values and the salient image information is detrimental to attribute editing, since the attributes represent the traits of face images. Therefore, rather than enforcing invariance of the latent representation to the attributes, \cite{he2017arbitrary} proposed an attribute classification constraint on the generated images, such that editing is wholly localised to the attributes one wishes to alter. Nevertheless, a principal issue with face editing is the lack of permanence with respect to the underlying identity of the original face image. Consequently, ~\cite{wang2018face} recently proposed an identity-preserving conditional generative adversarial network for face ageing synthesis, which ensures that the high-level representation of a synthesised aged face remains close to the representation of the original face. 

Profiting from these substantial advances, it was recently shown that MIIs could be synthesised~\cite{damer2019morgan} by following the approach of \cite{dosovitskiy2016generating}. However, \cite{scherhag2019face} conjectured that the attacks optimised in \cite{damer2019morgan}, for a specific face comparator, were unlikely to generalise to dissimilar face comparators.

\subsection{Input Attack Detection}
Unsurprisingly, the vulnerability of face comparators, in particular to image space face morphing, has fostered new research into targeted defences~\cite{seibold2017detection,asaad2017topological,ferrara2018face,debiasi2018prnu,raghavendra2016detecting}, which can be considered a specific problem within the field of general image tampering detection~\cite{birajdar2013digital}. Broadly speaking, the defences are either \emph{no-reference} or \emph{differential}~\cite{scherhag2019face}. No-reference methods process single images, e.g. an image submitted for enrolment to be stored; whereas, differential methods compare an image, captured by a trusted source, to its supposed corresponding stored database image.

Whilst we briefly review detection approaches below,\footnote{A comprehensive survey on face recognition systems under morphing attacks can be found in \cite{scherhag2019face}.} each proposed method suffers from the same underlying problem: they generalise poorly when the training and testing distributions differ~\cite{scherhag2018performance,spreeuwers2018towards}. In other words, they attempt to detect a specific cue using a supervised training approach; and if an attack avoids presenting this cue, then it will go undetected, even if the MIIs present other clear cues to their nature. 

\subsubsection{No-reference Detection}
Texture descriptors, based on hand-crafted shallow representations\footnote{For example, Local Binary Patterns (LBP), Binarised Statistical Image Features (BSIF), Scale-Invariant Feature Transform (SIFT), Speeded Up Robust Features (SURF), and Histogram of Oriented Gradients (HOG).} have been shown to be effective for detecting image tampering, particularly when combined with a supervised classifier~\cite{raghavendra2016detecting,scherhag2018towards,kraetzer2017modeling,asaad2017topological}. However, the descriptors used are typically of an extremely high dimensionality. Consequently, \cite{seibold2017detection} proposed deep supervised neural networks trained to detect face morphs, which result in more manageable and expressive representations. 

Amongst other approaches, \cite{debiasi2018prnu} analysed changes in noise patterns present with an image, since atypical changes imply that an image was captured by multiple devices---i.e. an image is likely to be the composite of multiple face images. Others have proposed to detect JPEG double-image compression~\cite{makrushin2017automatic,neubert2018extended}, premised on compression being an indicator of manipulation. In \cite{seibold2018reflection}, face morphing was found to affect the physical validity of skin illumination. 

\subsubsection{Differential Detection}
Far fewer works concentrate on differential detection. In \cite{ferrara2018face}, a de-morphing strategy was proposed, whereby a live image is \emph{subtracted} from its stored version, attempting to invert the morphing process, such that if the stored image is an MII, then the subtraction process will reveal two face images of two distinct individuals. However, the method depends heavily on the conditions in which the live image was captured~\cite{scherhag2019face}. Differently, \cite{scherhag2018detecting} compare landmark positions from live captured face images to stored versions, measuring the angles between corresponding aligned landmark position vectors. Nevertheless, due to high intra-class variation, the method cannot reliably detect signs of manipulation.

\section{Preliminaries}
\label{sec:prelim}
To begin, we introduce the datasets and face comparators that we experiment with, and end by formally defining a measure of MII attack success.

\subsection{Datasets}
We use three disjoint face image datasets, namely VGGFace2 (VGGF2)~\cite{cao2018vggface2}, Color FERET (C-FERET)~\cite{phillips1997feret}, and Flickr-Faces-HQ (FFHQ)~\cite{karras2018style}.

\textbf{VGGF2}~\cite{cao2018vggface2} is an unconstrained face image dataset, consisting of \num{\sim 3.3}M loosely cropped JPG images of \num{\sim 9.1}K persons, hence \num{\sim 360} face images per person. The face images were downloaded from Google Image Search, and vary in pose, illumination, size and quality. The intra-class variation of these images is representative of the variability that will be encountered by a verification system deployed in an unconstrained environment.

\textbf{C-FERET}~\cite{phillips1997feret} is a constrained high-quality face image dataset gathered in a semi-controlled environment, consisting of \num{\sim 11.3}K ${512\times768}$ labelled colour and grey PPM images of \num{994} persons (\num{\sim 12} images per person). The face images were collected over \num{15} sessions between 1993 and 1996. The images vary in pose (\num{13} unique yaw angles), eyewear, facial hair, hairstyle, and ethnicity. The intra-class variation of these images is much less than VGGF2.

\textbf{FFHQ}~\cite{karras2018style} is an unconstrained high-quality face image dataset, consisting of \num{70}K aligned and cropped ${1024\times1024}$ unlabelled colour PNG images. The images were downloaded from Flickr, and vary with regard to apparent age, hairstyle, facial hair, ethnicity, eyewear, headwear, apparel, and image background.

\subsection*{Pre-processing}
For all images, face alignment was performed using the dlib~\cite{king2009dlib} HOG-based face detector and a facial landmark predictor model~\cite{kazemi2014one}. The process retains only the most confident detection per face image. If a face was not detected then the image was discarded (\SI{\sim 11}{\percent}, \SI{\sim 4}{\percent} and \SI{\sim 1.4}{\percent} for VGGF2, C-FERET and FFHQ, respectively). For a detected face, the landmark predictor utilised an ensemble of regression trees to estimate the position of \num{68} landmarks, which are then used to transform each face to a canonical alignment. Alignment ensures that: (i) the faces are centred; (ii) the eye centroids lie on a horizontal line; and (iii) the faces are similar in scale. Next, we resized and rescaled the images such that they have a resolution of ${128\times128\times3}$ with pixel values in the range ${[0,1]}$. For C-FERET, we discarded all face images with a pose label corresponding to a nonzero yaw angle and any persons with fewer than two face images. For each person with more than two face images we only kept two. 

\begin{figure*}[!t]
\centering
\includegraphics[width=1\textwidth]{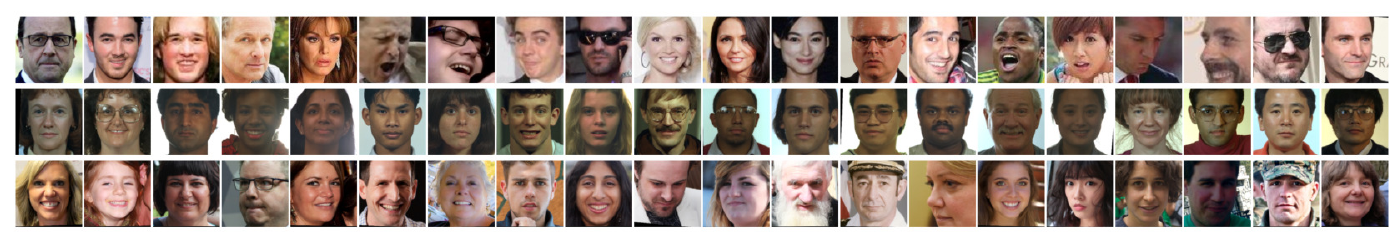}
\caption{Examples of the pre-processed aligned face images: VGGFace2 (top row); Color FERET (middle row); and FFHQ (bottom row).}
\label{fig:sample_preproc_ims}
\end{figure*}

\begin{figure}[!t]
\centering
\subfloat[VGGF2$_\text{eval}$]{\includegraphics[width=1\columnwidth]{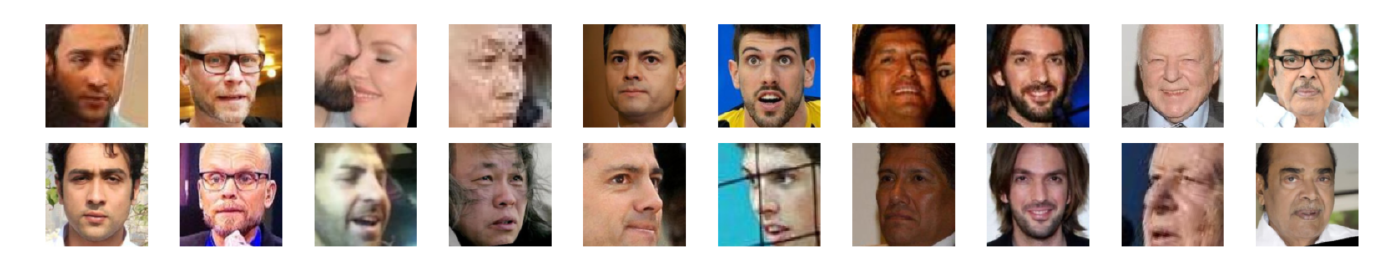}
\label{fig:var_a}}
\hfil
\subfloat[C-FERET]{\includegraphics[width=1\columnwidth]{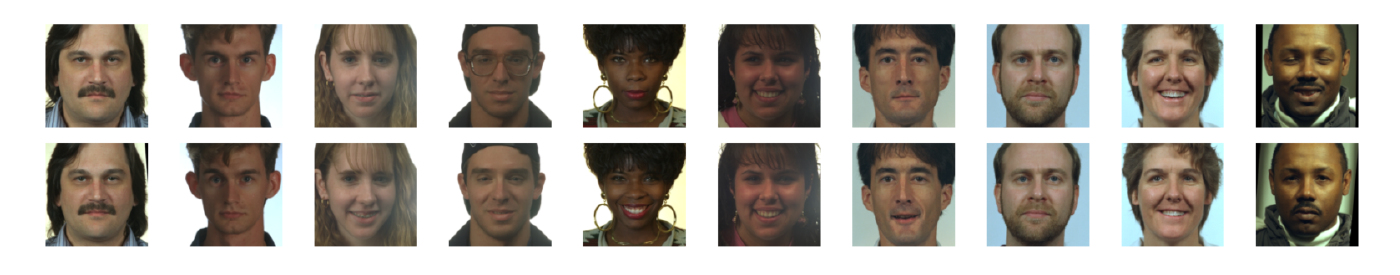}
\label{fig:var_b}}
\caption{Examples depicting the difference in intra-class variation between the VGGF2$_{\text{eval}}$ and C-FERET datasets. Each column in \Cref{fig:var_a,fig:var_b} shows two images (one considered a reference image and the other considered a live image) from a unique identity. It is clear that the two images, per identity, vary much more for VGGF2$_{\text{eval}}$ than for C-FERET.} \label{fig:var_dsets}
\end{figure}

In summary, we were left with: (i) \num{\sim 2.8}M VGGF2 train set partition (VGGF2$_\text{train}$) images; (ii) \num{\sim 152}K VGGF2 evaluation set partition (VGGF2$_\text{eval}$) images; (iii) \num{\sim 69}K FFHQ images; and (iv) \num{\sim 2}K C-FERET images of \num{993} persons. Examples of the pre-processed face images are shown in \cref{fig:sample_preproc_ims}; and \Cref{fig:var_dsets} compares the intra-class variability of VGGF2$_\text{eval}$ and C-FERET.

\subsection*{Dataset Usage}
Note that all MIIs were constructed for face image pairs sampled from the C-FERET dataset. We used VGGF2$_\text{eval}$ for face verification threshold determination, and VGGF2$_\text{train}$ for a gallery search attack (in \Cref{sec:real_im_attacks}), which utilises a real image (sampled from VGGF2$_\text{train}$) as an MII image. FFHQ was solely used by the representation space MII attack method for learning how to generate face images given a representation space encoding.

\subsection{Face Comparators}

In this work we employ three publicly available pre-trained softmax induced facial recognition neural networks: SENet$_{128}$~\cite{hu2018squeeze}, SENet$_{256}$~\cite{hu2018squeeze}, and LtNet$_{256}$~\cite{wu2018light}.\footnote{SENet$_{128}$ and SENet$_{256}$ are available at \url{https://github.com/ox-vgg/vgg_face2}, and are listed under the names SE-ResNet-50-128D and SE-ResNet-50-256D, respectively. LtNet$_{256}$ is available at \url{https://github.com/AlfredXiangWu/LightCNN}, and is listed under the name LightCNN-29v2.} All three networks output representations that can be used for face comparisons.

\textbf{SENet}~\cite{hu2018squeeze} integrates \emph{squeeze-and-excitation} (SE) blocks into a standard 50 layer residual network (ResNet)~\cite{he2016deep} architecture. SE units dynamically recalibrate channel-wise feature responses by explicitly modelling the relationships between channels, resulting in greater overall representational power. Both SENet$_{128}$ and SENet$_{256}$ were first pre-trained on the MS-Celeb-1M~\cite{guo2016ms} dataset, and then fine-tuned on the original VGGF2$_{\text{train}}$~\cite{cao2018vggface2} dataset of \num{8631} persons, using a softmax-based loss. At this point, SENet$_{128}$ and SENet$_{256}$ had the exact same weights.

A final stage of training was performed by projecting from the penultimate $2048D$ layer to either $128D$ (SENet$_{128}$) or $256D$ (SENet$_{256}$), and then fine-tuning the entire model by performing classification, i.e. $128\mapsto 8631$ or $256\mapsto 8631$. Hence SENet$_{128}$ and SENet$_{256}$ have different weights at every layer.

\textbf{LightCNN}~\cite{wu2018light} is a convolutional architecture consisting of 29 layers, utilising ResNet-inspired residual blocks, but using Max Feature Map (MFM) activation functions rather than ReLU~\cite{nair2010rectified} nonlinearities, implementing a variant of Maxout~\cite{goodfellow2013maxout}. MFM enforces a sparse relationship between consecutive layers by combining two feature maps and outputting their element-wise maximum. Moreover, no batch normalisation (BN) is used within the residual blocks. The penultimate layer containing the face representation is fully-connected with an MFM activation, as opposed to a global average pooling layer, which results in $256D$ representations. The network was trained on the CASIA-WebFace~\cite{yi2014learning} and MS-Celeb-1M~\cite{guo2016ms} datasets with a fully-connected classification layer over \num{80013} identities. We denote this model as LtNet$_{256}$. 

\subsubsection{Face Representations}
Consider a facial recognition neural network trained over a set $c$ of known identities. Ignoring the classification layer, one can take some intermediate layer as a representation of facial appearance. Following normal practice, we will use the layer immediately before the classification layer as giving an encoding of the input image into a representation space. Again following normal practice, we normalise the vector of outputs from this layer so that the network, parameterised by $\omega$, performs a mapping ${f_\omega:\mathcal{X} \rightarrow \mathbb{S}^{d-1}\subset \mathbb{R}^d}$; where $\mathcal{X}$ is the image space, $d$ is the number of units in the final layer, and $\mathbb{S}^{d-1}$ is the unit ${(d-1)}$-sphere. For compactness we denote the representation of an image $\mathbf{p}$ as ${\widetilde{\mathbf{p}}= f_\omega(\mathbf{p})}$.

\subsubsection{Representation Distances}
Given two face representations, $\widetilde{\mathbf{p}}$ and $\widetilde{\mathbf{q}}$, we use angular distance to quantify their \emph{visual} disparity, which for unit-length vectors is defined as:
\begin{align}\label{eq:ang_sim}\theta(\widetilde{\mathbf{p}},\widetilde{\mathbf{q}}) &= \arccos(\widetilde{\mathbf{p}} \cdot \widetilde{\mathbf{q}}) \in [0,\pi].
\end{align}

\subsubsection{Performance Metrics}
Let $\mathcal{P}^{+}$ and $\mathcal{P}^{-}$ denote the sets of face representation pairs ${(\widetilde{\mathbf{p}}, \widetilde{\mathbf{q}})}$ of matching (same identity) and non-matching (different identities), respectively. The evaluation of a face comparator is typically performed by determining a distance threshold $\epsilon$ for the binary classification of samples drawn from $\mathcal{P}^{+}$ and $\mathcal{P}^{-}$. 

The true acceptance rate (TAR) for $\epsilon$ is defined as the fraction of $\mathcal{P}^{+}$ that have a distance less than $\epsilon$. Conversely, the false acceptance rate (FAR) is defined as the fraction of $\mathcal{P}^{-}$ that have a distance less than $\epsilon$. A two-dimensional receiver operating characteristic (ROC) curve depicts the relationship between the FAR and TAR, as $\epsilon$ varies, with the area under the curve (AUROC) equal to the probability that a random ${(\widetilde{\mathbf{p}}, \widetilde{\mathbf{q}})\in\mathcal{P}^{+}}$ has a lower angular distance than a random ${(\widetilde{\mathbf{p}}, \widetilde{\mathbf{q}})\in\mathcal{P}^{-}}$. Furthermore, the \emph{strength} of a face comparator $f_\omega$ is typically assessed by computing the TAR at some specific FAR.

\subsubsection{Comparator Performance}
\label{sec:eps_threshold}
\begin{figure*}[!t]
\centering
\subfloat[SENet$_{128}$]{\includegraphics[width=0.3\textwidth]{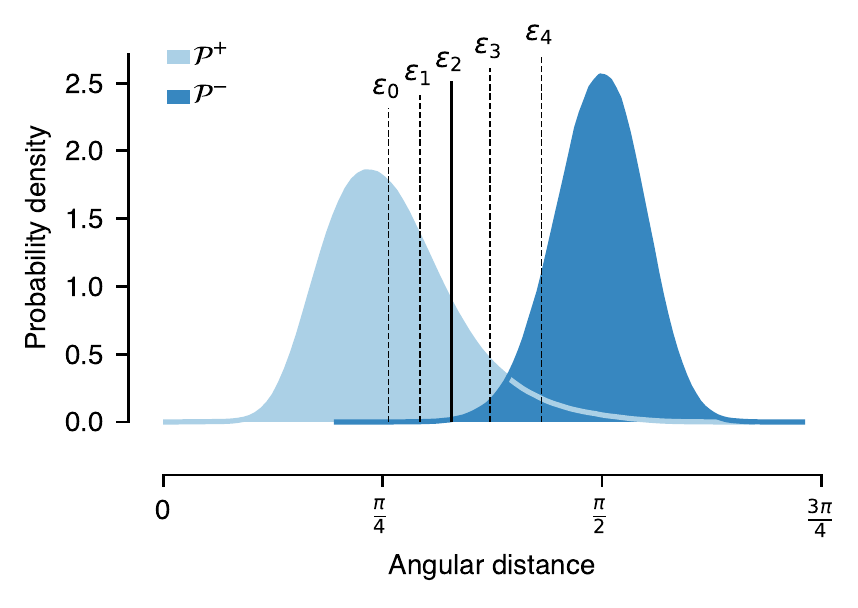}
\label{fig:match_sen}}
\hfil
\subfloat[SENet$_{256}$]{\includegraphics[width=0.3\textwidth]{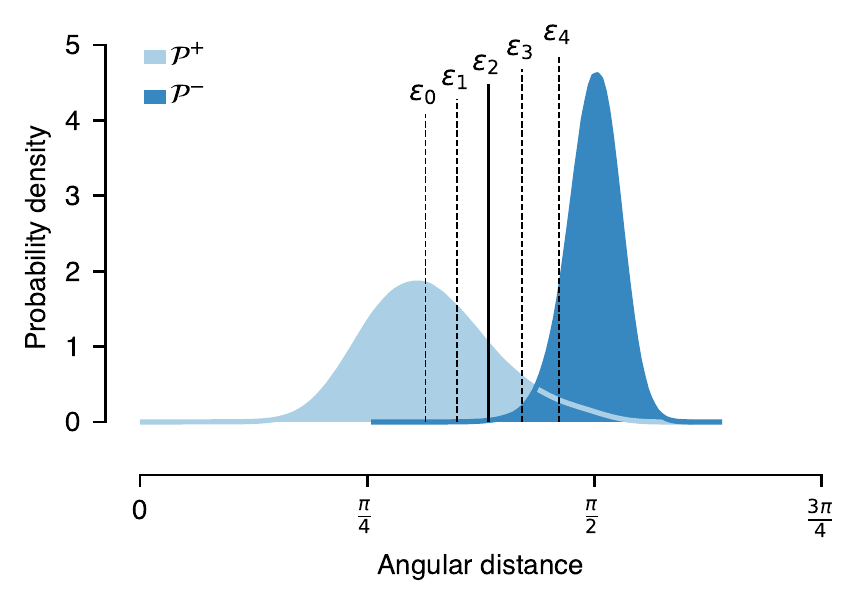}
\label{fig:match_sen256}}
\hfil
\subfloat[LtNet$_{256}$]{\includegraphics[width=0.3\textwidth]{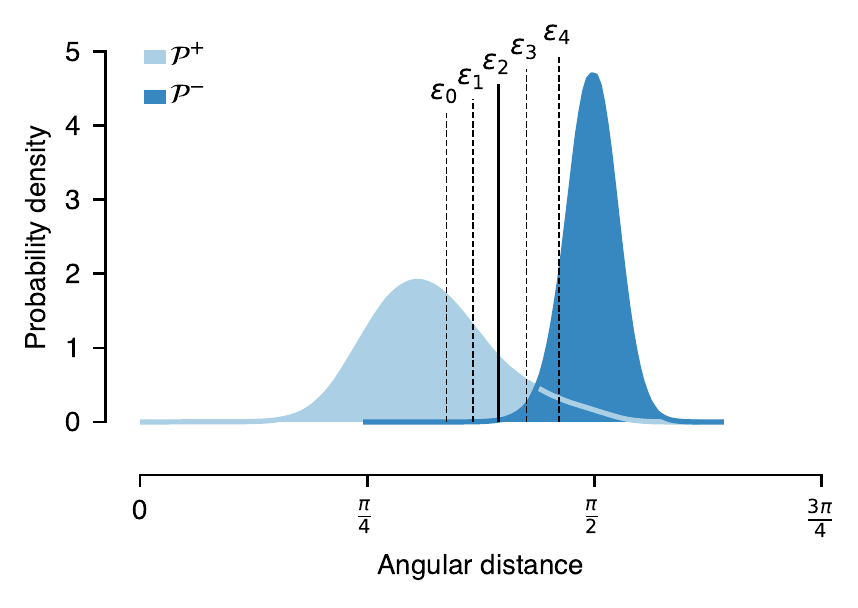}
\label{fig:match_light}}
\caption{Angular distance probability density plots of face representation pairs in $\mathcal{P}^{+}$ (pale blue) and $\mathcal{P}^{-}$ (dark blue), where $\mathcal{P}^{+}$ and $\mathcal{P}^{-}$ are sampled from VGGF2$_{\text{eval}}$. Overlaid are five different thresholds $\epsilon_i$, ${i=0,\dots,4}$ (vertical lines), based on the TAR at FAR ${p\in\{0.001\%,0.01\%\dots,10\%\}}$.} \label{fig:hists_cos_scores}
\end{figure*}

To assess the performance of SENet$_{128}$, SENet$_{256}$ and LtNet$_{256}$, we utilise the VGGF2$_\text{eval}$ dataset, which consists of faces of persons unused during training of any of the comparators. For threshold determination, we compute the TAR at FAR ${p\in\{0.001\%,0.01\%\dots,10\%\}}$---using all possible image pair combinations---which results in the set ${\{\epsilon_0,\epsilon_1,\dots, \epsilon_4\}}$. The probability density plots, in \Cref{fig:hists_cos_scores}, show the angular distance of pairs sampled from either $\mathcal{P}^{+}$ or $\mathcal{P}^{-}$, with the thresholds ${\{\epsilon_0,\epsilon_1,\dots, \epsilon_4\}}$ of each comparator overlaid. According to \Cref{table:correlation_coeffs}, the angular distances between pairs, in particular those sampled from $\mathcal{P}^{+}$, across the face comparators, exhibit a positive linear correlation. Thus, to a degree, each comparator is roughly performing in a similar manner; although, the arrangement of the face representations may differ from comparator to comparator.

\begin{table}[!t]
\centering
\begin{threeparttable}
\renewcommand{\arraystretch}{1.3}
\caption{Pearson correlation coefficients}
\label{table:correlation_coeffs}

\begin{tabularx}{\columnwidth}{ccYY}
\toprule
  & & SENet$_{256}$  &  LtNet$_{256}$\\ \midrule
SENet$_{128}$ & \begin{tabular}{@{}c@{}}$\mathcal{P}^{+}$ \\ $\mathcal{P}^{-}$\end{tabular}  & \begin{tabular}{@{}c@{}} $0.942$ \\ $0.691$\end{tabular}  & \begin{tabular}{@{}c@{}}$0.889$ \\ $0.478$\end{tabular} \\ \midrule
SENet$_{256}$ & \begin{tabular}{@{}c@{}}$\mathcal{P}^{+}$ \\ $\mathcal{P}^{-}$\end{tabular} &   & \begin{tabular}{@{}c@{}}$0.882$ \\ $0.489$\end{tabular}  \\ \bottomrule
\end{tabularx}
\begin{tablenotes}
\item Pearson correlation coefficients between angular distances according to the three face comparators.
\end{tablenotes}
\end{threeparttable}
\end{table}

\Cref{table:auroc_far} reports the performance, showing that the three comparators perform at similar levels. Following FRONTEX\footnote{European Agency for the Management of Operation Cooperation at the External Borders of the Member States of the European Union.} guidelines, it is recommended that deployed facial verification systems---in automated border control scenarios---should employ a threshold that gives a FAR of \SI{0.1}{\percent}, i.e. $\epsilon_2$. We will primarily focus on the $\epsilon_2$ threshold throughout the remainder of this work. At $\epsilon_2$, SENet$_{128}$, SENet$_{256}$ and LtNet$_{256}$ have a TAR of \SI{83.0}{\percent}, \SI{81.9}{\percent} and \SI{84.4}{\percent}, respectively, which implies that LtNet$_{256}$ is the strongest comparator. 

\begin{table}[!t]
\centering
\begin{threeparttable}
\renewcommand{\arraystretch}{1.3}
\caption{Performance summary of the face comparators}
\label{table:auroc_far}
\begin{tabularx}{\columnwidth}{lYYY>{\columncolor{black!5}}YYY}
\toprule
 \multirow{2}{*}{} &  \multirow{2}{*}{AUROC}  & \multicolumn{5}{c}{TAR at FAR $p$}    \\

 &   & \SI{0.001}{\percent} ($\epsilon_0$) & \SI{0.01}{\percent} ($\epsilon_1$) & \SI{0.1}{\percent} ($\epsilon_2$) & \SI{1}{\percent} ($\epsilon_3$) & \SI{10}{\percent} ($\epsilon_4$)\\
\midrule
SENet$_{128}$  & $\mathit{98.8}$  &  $52.8$ & $70.4$ & $83.0$ & $\mathit{91.8}$ & $\mathit{97.2}$ \\ 

SENet$_{256}$  & $98.4$  &  $49.7$ & $68.3$ & $81.9$ & $91.1$ & $96.5$ \\ 

LtNet$_{256}$ & $98.1$ & $\mathit{61.4}$ & $\mathit{75.2}$ & $\mathit{84.4}$ & $91.1$ & $95.8$  \\\bottomrule

\end{tabularx}
\begin{tablenotes}
\item Performance (in \%) of the three face comparators on the hold-out VGGF2$_\text{eval}$ set. Reported is the AUROC and TAR at FAR ${p\in\{0.001\%,0.01\%\dots,10\%\}}$. \emph{Italics} indicate the best performance per column. 
\end{tablenotes}
\end{threeparttable}
\end{table}

\subsection{Performance of a Multiple-Identity Image Attack}
\label{sec:mii_definition}
Consider an adversary ($p$) and accomplice ($q$), and images of them ${\mathbf{p}_{\text{ref}}, \mathbf{p}_{\text{live}}, \mathbf{q}_{\text{ref}}}$ and $\mathbf{q}_{\text{live}}$, where the \emph{reference} images are used to generate the MII, and the \emph{live} images are those captured at (for example) the \emph{entrance} to a facility which will be compared to the MII. Let ${h:\mathcal{X}^2\rightarrow \mathcal{X}}$ be a method to generate MIIs, so that ${h(\mathbf{p}_{\text{ref}}, \mathbf{q}_{\text{ref}})}$ is the attack image. The attack will be successful if and only if
\begin{align}
\theta( \widetilde{\mathbf{p}}_{\text{live}}, f_\omega(h(\mathbf{p}_{\text{ref}},\mathbf{q}_{\text{ref}}))) \leq \epsilon
\end{align}
and
\begin{align}
\theta( \widetilde{\mathbf{q}}_{\text{live}}, f_\omega(h(\mathbf{p}_{\text{ref}},\mathbf{q}_{\text{ref}}))) \leq \epsilon.
\end{align}
One condition representing the MII being accepted as a good likeness suitable for database storage, and the other representing identity verification at the facility. Using an MII angular distance defined as
\begin{align}
\widebar{\theta}((\widetilde{\mathbf{a}}, \widetilde{\mathbf{b}}), \widetilde{\mathbf{c}})=\max\{\theta(\widetilde{\mathbf{a}},\widetilde{\mathbf{c}}), \theta(\widetilde{\mathbf{b}},\widetilde{\mathbf{c}}) \},
\end{align}
the success condition can be rewritten as:
\begin{align}
\widebar{\theta}( (\widetilde{\mathbf{p}}_{\text{live}}, \widetilde{\mathbf{q}}_{\text{live}}), f_\omega(h(\mathbf{p}_{\text{ref}},\mathbf{q}_{\text{ref}}))) \leq \epsilon.
\end{align}
We define the overall success rate of $h$ as the fraction of successful attacks for randomly chosen pairs of distinct individuals.

\section{Why Are Multiple-Identity Images Possible?}
\label{sec:why_vul}
We now consider the behaviour and performance of a hypothetical \emph{ideal} method for MII generation. In later sections we will examine how well existing methods of MII generation realise this ideal. Breaking the problem in two like this will help in understanding why it is that MII attacks are possible.

The ideal MII attack method ${h^\star:\mathcal{X}^2\rightarrow\mathcal{X}}$ needs to minimise the MII distance
\begin{align}
\widebar{\theta}( (\widetilde{\mathbf{p}}_{\text{live}}, \widetilde{\mathbf{q}}_{\text{live}}), f_\omega(h^\star(\mathbf{p}_{\text{ref}},\mathbf{q}_{\text{ref}})))
\end{align}
to maximise its rate of success; and it must do this with reference only to $\mathbf{p}_{\text{ref}}$ and $\mathbf{q}_{\text{ref}}$, without direct knowledge of $\mathbf{p}_{\text{live}}$ and $\mathbf{q}_{\text{live}}$, which do not yet exist at the time the MII is generated. The best that can be done is to assume that ${\widetilde{\mathbf{p}}_{\text{live}}\approx\widetilde{\mathbf{p}}_{\text{ref}}}$ and ${\widetilde{\mathbf{q}}_{\text{live}}\approx\widetilde{\mathbf{q}}_{\text{ref}}}$ and thus aim to minimise the approximately equal MII distance
\begin{align}
\widebar{\theta}( (\widetilde{\mathbf{p}}_{\text{ref}}, \widetilde{\mathbf{q}}_{\text{ref}}), f_\omega(h^\star(\mathbf{p}_{\text{ref}},\mathbf{q}_{\text{ref}}))).
\end{align}
This proxy objective is easy to solve---$h^\star$ must generate an image whose representation is the spherical midpoint of the representations of the reference images, i.e. 
\begin{align}
\widetilde{\mathbf{m}}_{\text{ref}} &=f_\omega(h^\star(\mathbf{p}_{\text{ref}}, \mathbf{q}_{\text{ref}}))\nonumber\\
&= (\widetilde{\mathbf{p}}_{\text{ref}}+\widetilde{\mathbf{q}}_{\text{ref}})\lVert\widetilde{\mathbf{p}}_{\text{ref}}+\widetilde{\mathbf{q}}_{\text{ref}}\rVert_2^{-1},
\end{align}
which gives us:
\begin{align}\label{eq:thetabar}
\widebar{\theta}( (\widetilde{\mathbf{p}}_{\text{live}}, \widetilde{\mathbf{q}}_{\text{live}}), \widetilde{\mathbf{m}}_{\text{ref}}) &\approx \widebar{\theta}( (\widetilde{\mathbf{p}}_{\text{ref}}, \widetilde{\mathbf{q}}_{\text{ref}}), \widetilde{\mathbf{m}}_{\text{ref}}) \nonumber \\ 
&=\dfrac{1}{2}\theta(\widetilde{\mathbf{p}}_{\text{ref}}, \widetilde{\mathbf{q}}_{\text{ref}}).
\end{align}
Observe that the MII distances in this equation can be computed without actually implementing the ideal MII generator $h^\star$.

We have computed the MII distances for the ideal generator $h^\star$. In steps: for each identity $p$ we select distinct images $\mathbf{p}_{\text{ref}}$ and $\mathbf{p}_{\text{live}}$; we randomly pair individuals $p$ and $q$ to act as adversary and accomplice; for each pairing we compute $\widetilde{\mathbf{m}}_{\text{ref}}$, the representation of the ideal MII based on the reference images; using that we compute the ideal MII distance as ${\widebar{\theta}( (\widetilde{\mathbf{p}}_{\text{live}}, \widetilde{\mathbf{q}}_{\text{live}}), \widetilde{\mathbf{m}}_{\text{ref}})}$.

\begin{figure*}[!t]
\centering
\subfloat[SENet$_{128}$; VGGF2$_\text{eval}$]{\includegraphics[width=0.3\textwidth]{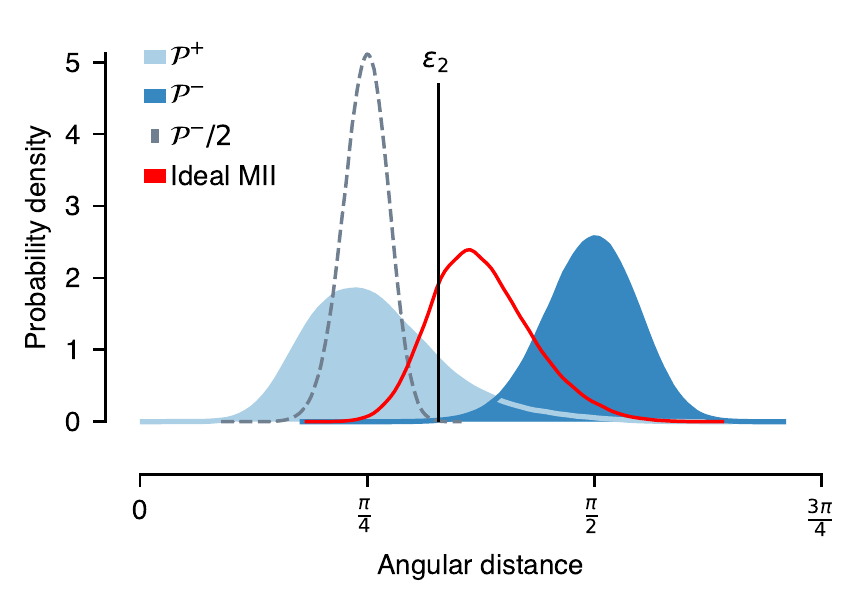}
\label{fig:ideal_match_sen}}
\hfil
\subfloat[SENet$_{256}$; VGGF2$_\text{eval}$]{\includegraphics[width=0.3\textwidth]{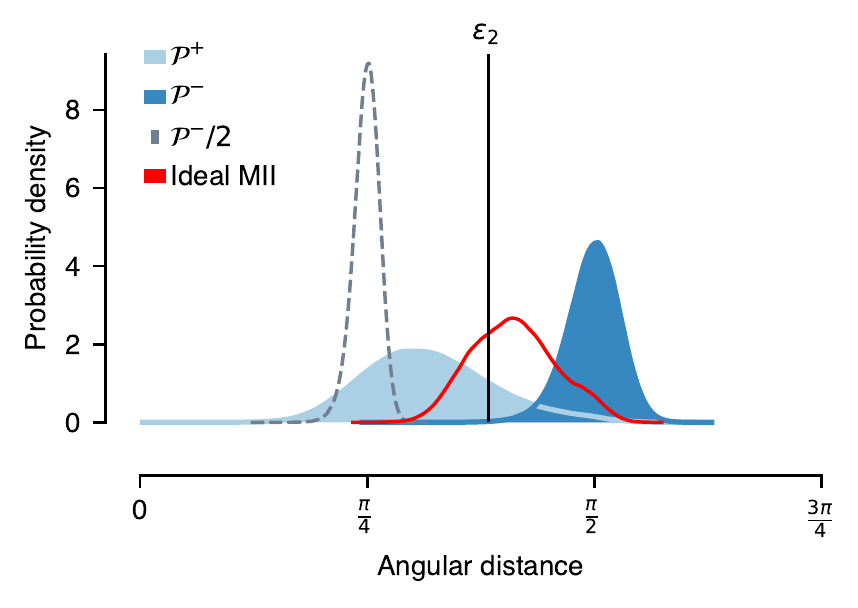}
\label{fig:ideal_match_sen256}}
\hfil
\subfloat[LtNet$_{256}$; VGGF2$_\text{eval}$]{\includegraphics[width=0.3\textwidth]{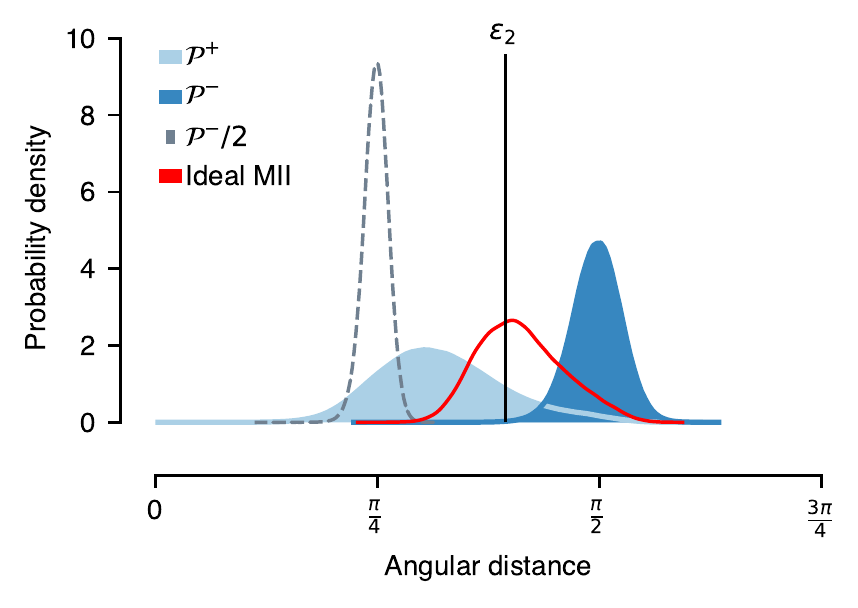}
\label{fig:ideal_match_light}}
\hfil
\subfloat[SENet$_{128}$; C-FERET]{\includegraphics[width=0.3\textwidth]{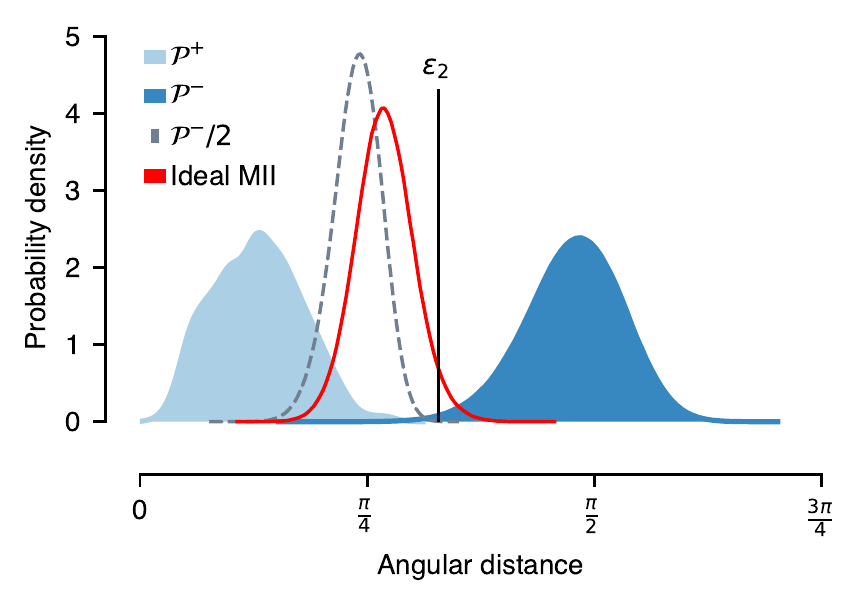}
\label{fig:ideal_match_sen_feret}}
\hfil
\subfloat[SENet$_{256}$; C-FERET]{\includegraphics[width=0.3\textwidth]{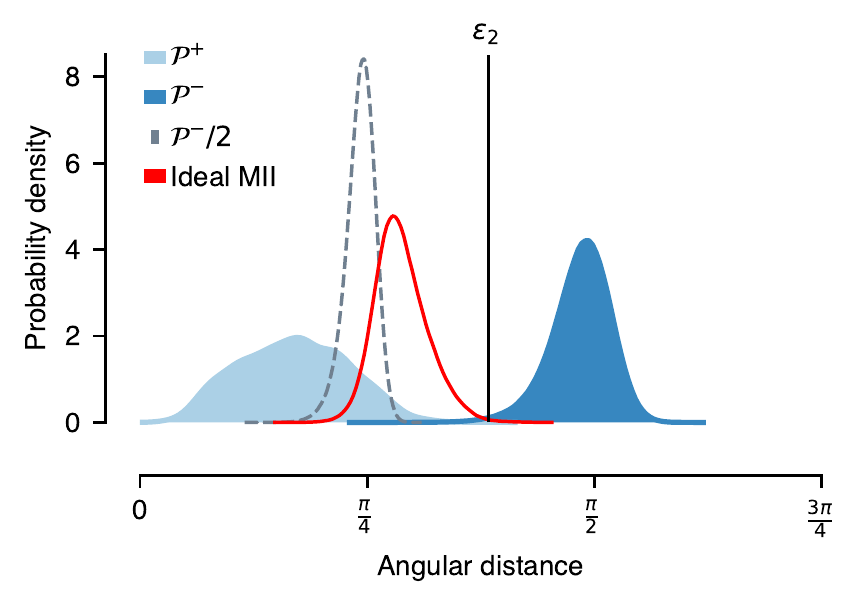}
\label{fig:ideal_match_sen256_feret}}
\hfil
\subfloat[LtNet$_{256}$; C-FERET]{\includegraphics[width=0.3\textwidth]{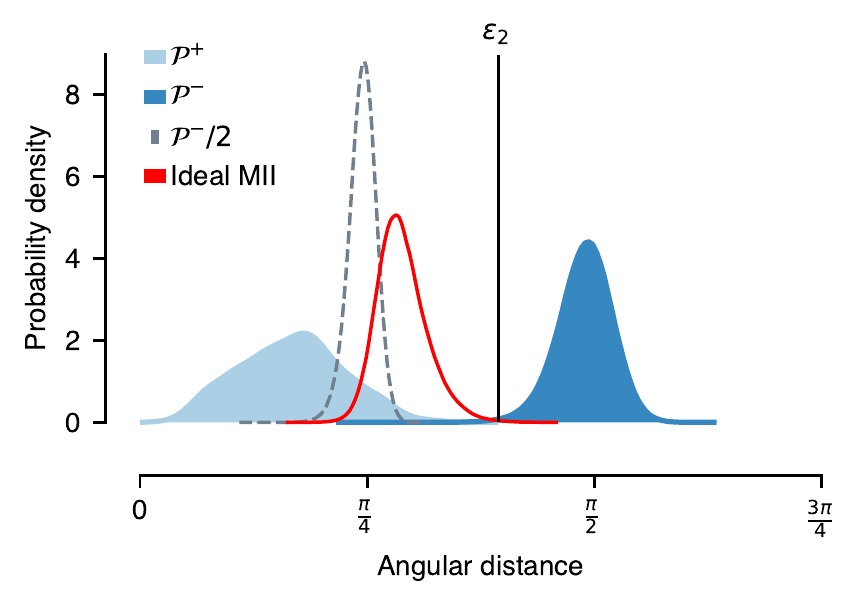}
\label{fig:ideal_match_light_feret}}
\caption{Angular distance probability density plots of MII distances for an ideal attack method, i.e. $\widebar{\theta}( (\widetilde{\mathbf{p}}_{\text{live}}, \widetilde{\mathbf{q}}_{\text{live}}), \widetilde{\mathbf{m}}_{\text{ref}})$ (red); the distributions of angular distances between $\mathcal{P}^{+}$ (pale blue) and $\mathcal{P}^{-}$ (dark blue) pairs; and the distribution of $\mathcal{P}^{-}$ paired angular distances divided by two (corresponding to the case when $\widetilde{\mathbf{p}}_{\text{live}}=\widetilde{\mathbf{p}}_{\text{ref}}$ and $\widetilde{\mathbf{q}}_{\text{live}}=\widetilde{\mathbf{q}}_{\text{ref}}$), denoted as $\mathcal{P}^{-}/2$ (dashed grey). \Cref{fig:ideal_match_sen,fig:ideal_match_sen256,fig:ideal_match_light} use the unconstrained VGGF2$_{\text{eval}}$; and \Cref{fig:ideal_match_sen_feret,fig:ideal_match_sen256_feret,fig:ideal_match_light_feret} use the constrained C-FERET dataset of frontally aligned images. Overlaid is the $\epsilon_2$ threshold (vertical line) based on the TAR at FAR $p=0.1\%$ computed using matching and non-matching pairs sampled from VGGF2$_{\text{eval}}$, which is the threshold used for all plots.} \label{fig:ideal_hists_cos_scores}
\end{figure*}

The distribution of ideal MII distances is shown in \Cref{fig:ideal_hists_cos_scores} for attacks against all three face comparators, and when the reference and live images are drawn either from VGGF2$_\text{eval}$ or from C-FERET. Each plot also shows the distribution of angular distances between $\mathcal{P}^{+}$ and $\mathcal{P}^{-}$ pairs from the same dataset. Additionally the plots show the distribution of half $\mathcal{P}^{-}$ distances, which from \Cref{eq:thetabar} we expect to approximate the MII distances. Finally, the plots show the $\epsilon_2$ threshold that MII distances need to be below for an attack to succeed. There is much to see in \Cref{fig:ideal_hists_cos_scores}:
\begin{itemize}
    \item In all plots the ${\mathcal{P}^{-}/2}$ distances (dashed) are fully below the $\epsilon_2$ threshold. Thus if an attack could achieve these distances it would always be successful.
    \item The ideal MII distances (red) are always greater than ${\mathcal{P}^{-}/2}$ distances, since the MIIs are at the midpoints of the reference images, but are compared to the live images.
    \item For C-FERET images, for which intra-class variation is small, the MII distances are only slightly larger than the ${\mathcal{P}^{-}/2}$ distances, so the great majority (${{\sim}95\%}$) of ideal attacks would succeed.
    \item For VGGF2$_\text{eval}$ images, for which intra-class variation is larger, the MII distances are substantially larger than ${\mathcal{P}^{-}/2}$ distances, but still much less than $\mathcal{P}^{-}$ distances; resulting in attack success rates of ${{\sim}40\%}$.
\end{itemize}

In summary: if an adversary can ensure that reference and live images are very similar, then MII attacks will very likely succeed if the generator used is near ideal. The attack exploits the fact that a comparator computes distances between face images that are a compound of \emph{real} face differences and \emph{incidental} differences due to ageing, hairstyle, etc. An adversary can take advantage of this by constructing an MII which minimises incidental differences from its reference images \emph{and} whose real difference from its reference images is only half the normal distance for a pair of mismatched faces.

It remains to be explained why an MII that is half the normal mismatch distance from its reference images is so clearly below the threshold distance; equivalently, how come $\mathcal{P}^{-}$ distances are mostly above the threshold and ${\mathcal{P}^{-}/2}$ mostly below? The answer is that the $\mathcal{P}^{-}$ distribution is tight, and the position of the $\epsilon$ threshold (determined by the desired FAR) is necessarily in its left tail. Change either fact and MIIs would not succeed as often.

On the narrowness of the $\mathcal{P}^{-}$ distance distribution, observe that it is centred around $\pi/2$, the angular distance for orthogonal points. This indicates that face representations are very widely distributed on the sphere $\mathbb{S}^{d-1}$. Indeed, it can be shown that, for points uniformly distributed on $\mathbb{S}^{d-1}$, the angular distances are normally distributed with mean $\pi/2$ and standard deviation $d^{-1}$. The standard deviations of the $\mathcal{P}^{-}$ distances for the three comparators are slightly larger, i.e. ${\sim}2d^{-1}$, indicating that face representation are not fully uniformly distributed over $\mathbb{S}^{d-1}$, but are sufficiently uniform to give the tight distribution observed.

On the position of the $\epsilon$ threshold, observe that setting it lower would substantially impact the TAR rate, leading to unusable systems. This is the case because the comparators have achieved an acceptable FAR/TAR trade-off with $\mathcal{P}^{+}$ distances that are only modestly below orthogonal, rather than near zero. In plain terms, the current generation of comparators work because the representations of different identity faces are always close to orthogonal, and the representations of same identity faces are only slightly closer than orthogonal. This leaves them vulnerable to attack as an MII image can be much closer to its constituent faces than orthogonal.

\section{Generating Multiple-Identity Images}
\label{sec:constructing_miis_methods}
Here we outline three methods for either \emph{finding}, \emph{constructing}, or \emph{synthesising} MIIs. Our aim is to show that real MIIs can be generated that are sufficiently close to ideal MIIs that they are effective attacks.

\subsection{Finding Multiple-Identity Images by Gallery Search}
\label{sec:real_im_attacks}
The simplest form of attack uses a real face image as MII, chosen from a gallery to be as close as possible to both reference images. 

Formally, let ${(\mathbf{p}_{\text{ref}}, \mathbf{q}_{\text{ref}}) \in \mathcal{P}^{-}}$ be the reference images of the adversary and accomplice, and let $\mathcal{G}$ be a gallery of face images that they have access to. We define the gallery search MII (GS-MII) generator $h_\text{GS}$ as picking the ${\mathbf{m}_{\text{ref}}=h_\text{GS}(\mathbf{p}_{\text{ref}}, \mathbf{q}_{\text{ref}})}$ in $\mathcal{G}$ such that ${\widebar{\theta}( (\widetilde{\mathbf{p}}_{\text{ref}}, \widetilde{\mathbf{q}}_{\text{ref}}), \widetilde{\mathbf{m}}_{\text{ref}})}$ is minimised, i.e. the gallery image whose representation is as close as possible to the reference representations. Note that this attack makes use of representation distances, so the adversary needs access to a comparator to compute them---and this comparator may be the same or different to the one attacked.

\subsection{Constructing Multiple-Identity Images in Image Space}
\label{sec:im_space_attacks}
Face morphing is the image space process of geometric warping and then colour interpolating two distinct face images into a single composite~\cite{ferrara2014magic,scherhag2018performance}.

We use the following, standard, process to construct image space MIIs (IS-MIIs). Given a pair of frontally aligned face images, say ${(\mathbf{p}_{\text{ref}},\mathbf{q}_{\text{ref}})\in\mathcal{P}^{-}}$, an IS-MII is constructed as follows: (i) corresponding facial landmark position vectors are determined in $\mathbf{p}_{\text{ref}}$ and $\mathbf{q}_{\text{ref}}$ using the dlib landmark detector~\cite{kazemi2014one}\footnote{The detector results in \num{68} facial landmarks; in addition, we define \num{20} evenly spaced landmarks on the image boundary.}; (ii) Delaunay triangulation~\cite{delaunay1934sphere} is performed on the average of the two sets of landmark position vectors; (iii) an affine transformation is applied to the image within each triangle to transform $\mathbf{p}_{\text{ref}}$ and $\mathbf{q}_{\text{ref}}$ to the averaged positions, which results in two warped images; and (iv) the two warped images are averaged to give an IS-MII ${\mathbf{m}_{\text{ref}}= h_{\text{IS}}(\mathbf{p}_{\text{ref}},\mathbf{q}_{\text{ref}})}$. Note that this method for generating MIIs does not require use of a comparator.

\subsection{Synthesising Multiple-Identity Images via Representation Space}
\label{sec:deep_space_attacks}
Assuming a face comparator has learnt disentangled high-level representations that encode the identity of a face image, then in principle we can learn an inverse mapping. That is, we can learn to generate face images from abstract face representations. Having learnt such a mapping, we can synthesise MIIs by first constructing the ideal midpoint of the reference image representations, and then synthesising a face image corresponding to that midpoint.

Concisely, let $g_\phi$ be a deconvolutional decoder, i.e. ${g_\phi:\mathbb{R}^d\rightarrow \mathcal{X}}$, parameterised by $\phi$, and $\widetilde{\mathbf{m}}_{\text{ref}}$ denote the spherical midpoint of $\widetilde{\mathbf{p}}_{\text{ref}}$ and $\widetilde{\mathbf{q}}_{\text{ref}}$, where ${(\mathbf{p}_{\text{ref}},\mathbf{q}_{\text{ref}})\in\mathcal{P}^{-}}$. We define a synthesised MII as ${\mathbf{m}_{\text{ref}}=h_{\text{RS}}(\mathbf{p}_{\text{ref}},\mathbf{q}_{\text{ref}})=g_\phi(\widetilde{\mathbf{m}}_{\text{ref}})}$ and will refer to it as a representation space MII (RS-MII). 

\subsubsection{Model}
To this end, we learn $g_\phi$, which is tasked with reproducing the face image ${\mathbf{p}\in\mathcal{X}}$ given its representation ${\widetilde{\mathbf{p}}=f_\omega(\mathbf{p})\in\mathbb{R}^d}$, where $f_\omega$ is a face comparator with fixed weights. Inspired by \cite{dosovitskiy2016generating}, we learn $g_\phi$ using a combination of loss functions:
\begin{align}\label{eq:total_loss_gan}
\mathcal{L}&= \lambda_{\text{pix}}  \mathcal{L}_1^{\text{pix}} + \lambda_{\text{adv}}  \mathcal{L}^{\text{adv}} + \lambda_{\text{feat}}  \mathcal{L}_2^{\text{feat}},
\end{align}
where ${\lambda_{\text{pix}}, \lambda_{\text{feat}}, \lambda_{\text{adv}}\in\mathbb{R}_{\geq 0}}$ are weights. For the autoencoding pixel-wise loss $\mathcal{L}_1^{\text{pixel}}$, we utilise the mean absolute deviations:
\begin{align}
\label{fig:loss_l1}
\mathcal{L}_1^{\text{pixel}} = \frac{1}{n} \sum_{i=1}^{n}\lVert g_\phi(\widetilde{\mathbf{p}}) - \mathbf{p} \rVert_1.
\end{align}

Whilst the $\mathcal{L}_1^{\text{pix}}$ loss enforces low-frequency correctness, it generally fails to produce satisfactory results in terms of high-frequency features. Therefore, we also learn an image patch convolutional discriminator (PatchGAN)~\cite{isola2017image} $d_{\psi}$, parameterised by $\psi$, which attends to localised high-frequency details. The PatchGAN moves across an image, classifying patches as being either real or reconstructed. The discriminator and the decoder are trained simultaneously. This process forces the decoder to attempt to produce reconstructions indistinguishable from real data. Concretely, the discriminator parameters $\psi$ are learnt by minimising a least squares loss~\cite{mao2017least}:
\begin{align}
\label{fig:loss_discr}
\mathcal{L}^{\text{disc}} = \frac{1}{n} \sum_{i=1}^{n} [d_\psi(\mathbf{p})-1]^2 + [d_\psi(g_\phi(\widetilde{\mathbf{p}}))]^2,
\end{align}
with the decoder trained to minimise:
\begin{align}
\label{fig:loss_adv}
\mathcal{L}^{\text{adv}} = \frac{1}{n} \sum_{i=1}^{n}[d_\psi(g_\phi(\widetilde{\mathbf{p}}))-1]^2. 
\end{align}

Lastly, the feature loss $\mathcal{L}_2^{\text{feat}}$ ensures that the representation of a decoded image matches the representation of the original image:
\begin{align}
\label{fig:loss_feat}
\mathcal{L}_2^{\text{feat}} = \frac{1}{n} \sum_{i=1}^{n}\lVert f_\omega(g_\phi(\widetilde{\mathbf{p}})) - \widetilde{\mathbf{p}}  \rVert_2^{2}.
\end{align}

\subsubsection{Implementation}
The encoder network $f_\omega$ is fixed and is defined as SENet$_{128}$. 

Let $c_k$ denote a ${4\times4}$ ConvTranspose-BN-ReLU layer with $k$ filters, stride $2$ and padding $1$ followed by a ${3\times3}$ Conv-BN-ReLU layer with $k$ filters, stride $1$ and padding $1$. The architecture of the decoder $g_\phi$ is:
\begin{align}\nonumber
c_{512} - c_{256} - c_{128} - c_{64} - c_{32} - c_{16}.
\end{align}
After the last layer, a Conv-Tanh layer is applied, producing a $3$-channel output. The output of $g_\phi$ given a ${1\times128}$ sized input is a ${128\times128\times3}$ image.

With respect to the discriminator $d_\psi$, we utilise a ${70\times70}$ PatchGAN. Let $c_k$ be a ${4\times4}$ Conv-BN-LeakyReLU layer with $k$ filters, stride $2$ and padding $1$. The slope of the LeakyReLU activations is equal to $0.2$. The discriminator architecture $d_\psi$ is:
\begin{align}\nonumber
c_{64} - c_{128} - c_{256} - c_{512}.
\end{align}
The discriminator accepts inputs of size ${128\times128\times3}$.

\subsubsection{Model Training}
The weights used in \Cref{eq:total_loss_gan} were empirically set as follows: $\lambda_{\text{pix}}=10$, $\lambda_{\text{feat}}=300$, and $\lambda_{\text{adv}}=1$. Training was performed on the FFHQ dataset (\SI{90}{\percent} for training and \SI{10}{\percent} for validation). The training set was augmented by on-the-fly horizontal mirroring. For pre-processing, with respect to the discriminator, the image pixels were all normalised to the range $[-1,1]$. The Adam solver was used with a fixed learning rate of \num{2e-4}, $\beta_1=0.5$, and a batch size of \num{32}. The model was trained for \num{1000} epochs (\num{50000} iterations per epoch).

To stabilise training, the discriminator was updated using a history of \num{500} previously reconstructed images, as opposed to only the most recent. For inputs to the discriminator, we also used additive Gaussian white noise, with ${\mu=0}$ and ${\sigma=0.1}$, which we linearly decayed to zero over the first \num{600} epochs.

\subsection{Multiple-Identity Image Methods: A Comparison}
\label{sec:miis_comparison}
There are certain caveats associated with each MII attack method.
\begin{itemize}
    \item \textbf{GS-MIIs} rely on access to a gallery $\mathcal{G}$ and a face comparator $f_\omega$, in order to find the \emph{best} ${\mathbf{m}_{\text{ref}}\in\mathcal{G}}$ for a pair ${(\mathbf{p}_{\text{ref}}, \mathbf{q}_{\text{ref}})}$. For evaluation, we define $\mathcal{G}$ as the entire VGGF2$_{\text{train}}$ set of \num{\sim2.8}M face images, and assume that an adversary has access to the SENet$_{128}$ comparator.
    \item \textbf{IS-MIIs} require frontally aligned face images, such that corresponding landmarks between $\mathbf{p}_{\text{ref}}$ and $\mathbf{q}_{\text{ref}}$ can be found.
    \item \textbf{RS-MIIs} depend on having access to a face comparator $f_\omega$, so as to train a decoder $g_\phi$. In our evaluation we fix that the adversary has access to the SENet$_{128}$ comparator for this purpose.
\end{itemize}

\begin{figure*}[!t]
\centering
\includegraphics[width=1\textwidth,clip=true, trim = 0 15 0 0]{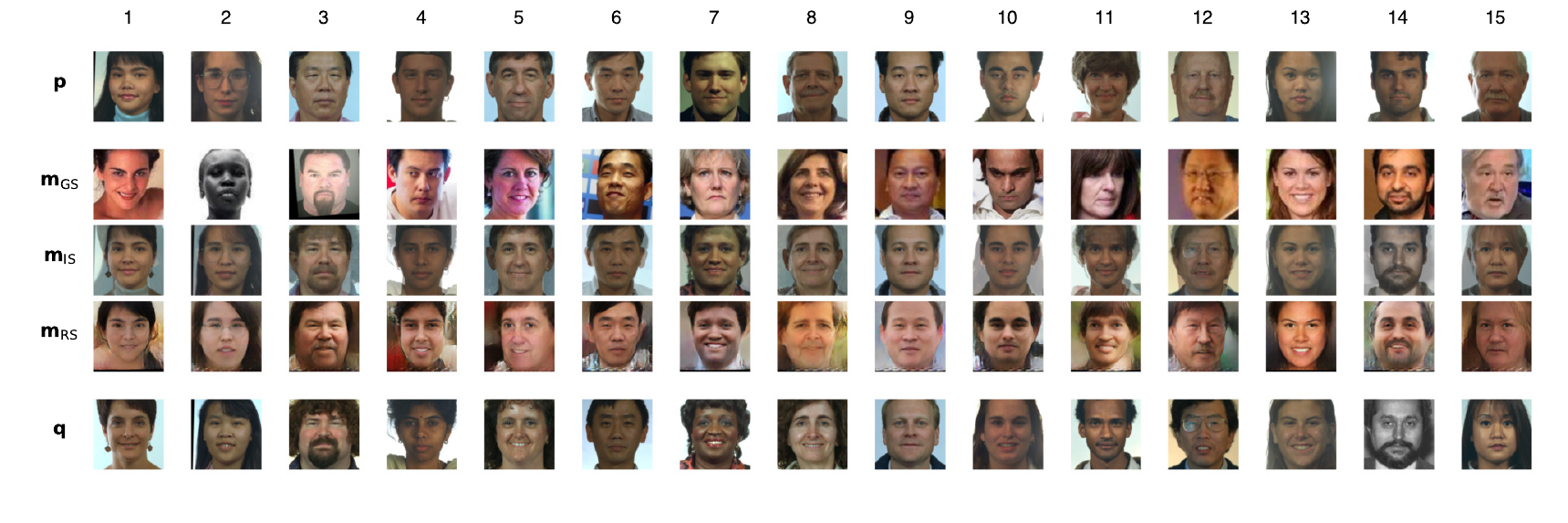}
\caption{Examples of random GS-MII (second row), IS-MII (third row) and RS-MII (fourth row) attacks given two non-matching images $\mathbf{p}$ (top row) and $\mathbf{q}$ (bottom row). Each column denotes a unique attack given $\mathbf{p}$ and $\mathbf{q}$.}
\label{fig:methods_comp}
\end{figure*}

In \Cref{fig:methods_comp}, we show examples of MIIs generated using the previously described methods, where $\mathbf{p}$ and $\mathbf{q}$ are sampled from the C-FERET dataset of frontally aligned face images. Qualitatively speaking, the GS-MIIs tend to resemble $\mathbf{p}$ and $\mathbf{q}$ at only a rudimentary level; and are sometimes far from satisfactory, in particular when $\mathbf{p}$ and $\mathbf{q}$ differ in ethnicity (e.g. column 2), gender (e.g. columns 4, 5 and 15) or both (e.g. columns 7, 11 and 12). The overarching benefit of GS-MIIs is that they are real images, therefore free from visual artefacts. In contrast, since IS-MIIs and RS-MIIs \emph{combine} $\mathbf{p}$ and $\mathbf{q}$ when constructing $\mathbf{m}$, the MIIs have much better visual similarity to both $\mathbf{p}$ and $\mathbf{q}$. However, IS-MIIs suffer from visual ghosting artefacts, and RS-MIIs lack detail at the transition between the face and background.

\section{Effectiveness of Multiple-Identity Images}
\label{sec:effectiveness}
Here we evaluate whether generated MIIs are close enough to ideal MIIs to make effective attacks. We also evaluate how much the effectiveness of generated MIIs depends on whether the comparator that is being attacked is the same as the one used to generate the MIIs.

\subsection{Experiment}
We split the C-FERET dataset of frontally aligned images into two disjoint sets (reference and live); each consisting of 993 face images of 993 unique persons. Each method for creating MIIs is applied to the same \num{10}K randomly sampled unique pairings $(\mathbf{p}_{\text{ref}},\mathbf{q}_{\text{ref}})$ from the reference set. We score the effectiveness of a generated MII $h(\mathbf{p}_{\text{ref}},\mathbf{q}_{\text{ref}})$ by the MII distance, i.e. 
\begin{align}\label{eq:probe}
\widebar{\theta}( (\widetilde{\mathbf{p}}_{\text{live}}, \widetilde{\mathbf{q}}_{\text{live}}), f_\omega(h(\mathbf{p}_{\text{ref}},\mathbf{q}_{\text{ref}}))).
\end{align}
The overall score for an MII generator $h$ is the fraction of these distances which are less than or equal to some $\epsilon$ threshold.

GS-MIIs and RS-MIIs require a comparator for their operation and we fix that to be SENet$_{128}$; IS-MIIs do not make use of a comparator. We evaluate the MIIs as attacks against SENet$_{128}$ (matched comparator mode), SENet$_{256}$ (similar comparator mode), and LtNet$_{256}$ (mismatched comparator mode).

\subsection{Analysis and Results}
\begin{figure*}[!t]
\centering
\subfloat[SENet$_{128}$]{\includegraphics[width=0.3\textwidth]{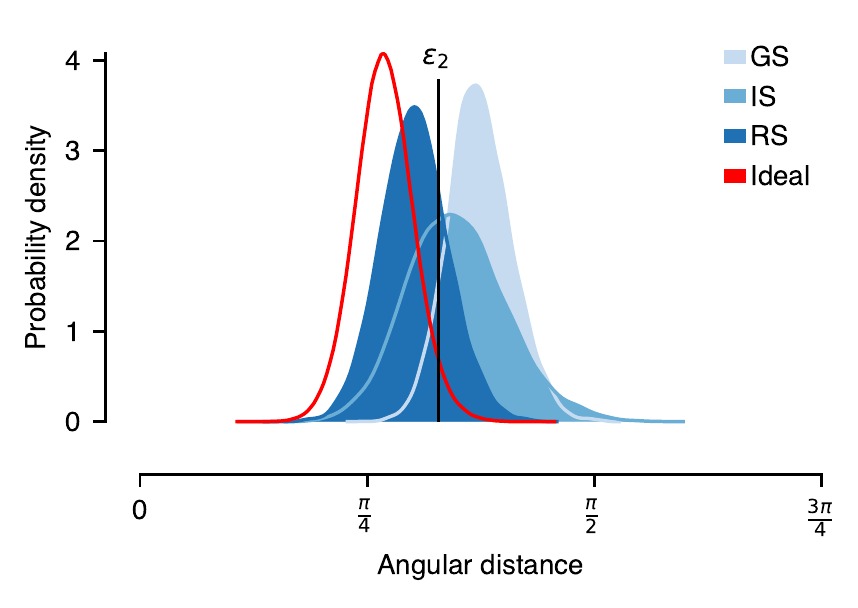}
\label{fig:succ_sen}}
\hfil
\subfloat[SENet$_{256}$]{\includegraphics[width=0.3\textwidth]{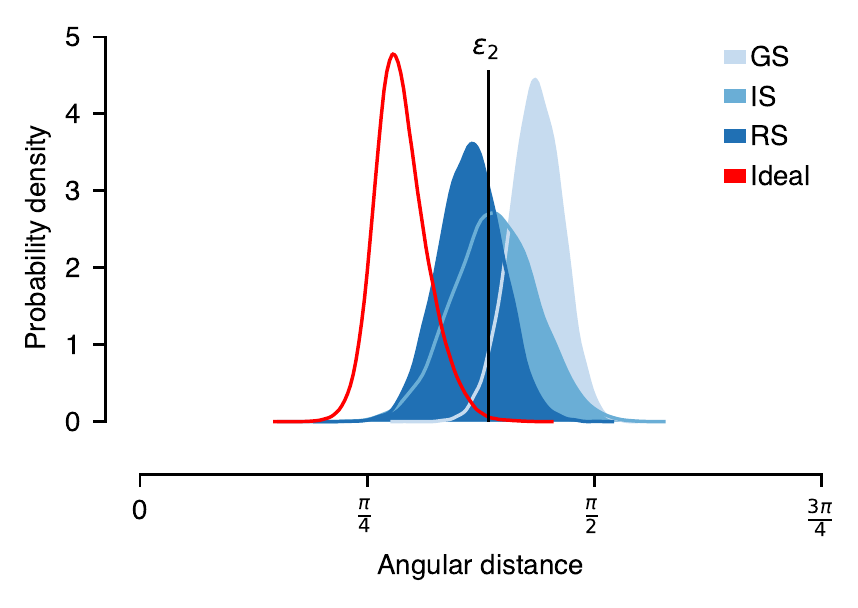}
\label{fig:succ_sen256}}
\hfil
\subfloat[LtNet$_{256}$]{\includegraphics[width=0.3\textwidth]{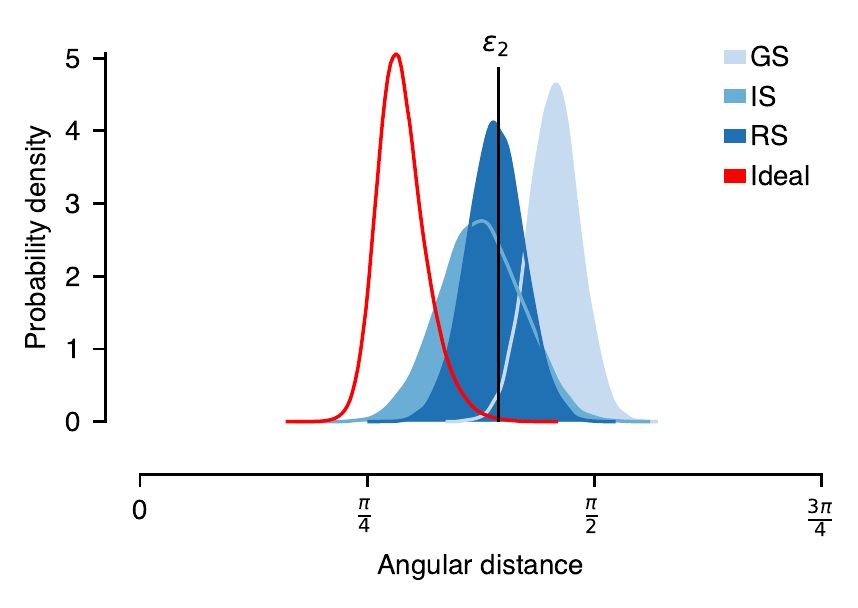}
\label{fig:succ_light}}
\caption{Angular distance probability density plots of \Cref{eq:probe} for GS-MII (pale blue), IS-MII (medium blue), RS-MII (dark blue), as well as the MIIs distances for an ideal attack method $h^\star$ (red). Overlaid is the $\epsilon_2$ threshold (vertical line) based on the TAR at FAR $p=0.1\%$.}  \label{fig:succ_cos_scores}
\end{figure*}

In \Cref{fig:succ_cos_scores}, we compare the distributions of MII distances for MIIs generated using the different methods, as well as the effect on the distributions as we vary the comparator attacked. The figure also shows the distributions of MII distances for ideal MIIs (as discussed in \Cref{sec:why_vul}). Clearly all forms of generated MII fail to realise the optimal MII attack, but the large amount of mass still below the $\epsilon_2$ threshold indicates that they are successful more often that not. \Cref{table:succ_comps} gives the exact MII success rates for a range of $\epsilon$ values, but we concentrate on $\epsilon_2$ corresponding to a FAR of 0.1\% for normal data.

The weak performance of GS-MIIs is evident and expected: in matched comparator mode only ${\sim}9\%$ of MII attacks succeed, and in similar and mismatched mode this drops below 4\%. 

\begin{table}[!t]
\centering
\begin{threeparttable}
\renewcommand{\arraystretch}{1.3}
\caption{Performance summary of the MII attacks}
\label{table:succ_comps}
\begin{tabularx}{\columnwidth}{YYYY>{\columncolor{black!5}}YYY}
\toprule

 & MII & \SI{0.001}{\percent} ($\epsilon_0$) & \SI{0.01}{\percent} ($\epsilon_1$) & \SI{0.1}{\percent} ($\epsilon_2$) & \SI{1}{\percent} ($\epsilon_3$) & \SI{10}{\percent} ($\epsilon_4$)\\
\midrule
\multirow{3}{*}{\scriptsize{$\text{SENet}_{128}$}}  & GS & $0.0$ & $0.7$ & $8.6$ & $49.9$ & $94.3$ \\ 
& IS &  $3.6$ & $14.8$ & $37.2$ & $68.3$ & $93.0$ \\ 
& RS &  $\mathit{10.6}$ & $\mathit{39.7}$ & $\mathit{76.9}$ & $\mathit{96.7}$ & $\mathit{100.0}$ \\  \midrule

\multirow{3}{*}{\scriptsize{$\text{SENet}_{256}$}}  & GS &  $0.0$ & $0.2$ & $3.8$ & $29.5$ & $81.5$ \\ 
& IS &  $4.9$ & $18.9$ & $44.0$ & $74.1$ & $93.7$ \\ 
& RS &  $\mathit{8.0}$ & $\mathit{34.0}$ & $\mathit{71.0}$ & $\mathit{95.3}$ & $\mathit{99.7}$ \\  \midrule

\multirow{3}{*}{\scriptsize{LtNet$_{256}$}}   & GS &  $0.0$ & $0.1$ & $1.4$ & $13.5$ & $57.6$ \\ 
& IS &  $\mathit{19.7}$ & $\mathit{42.8}$ & $\mathit{66.4}$ & $85.7$ & $97.0$ \\ 
& RS &  $4.4$ & $21.2$ & $54.2$ & $\mathit{86.8}$ & $\mathit{98.8}$ \\ \bottomrule
\end{tabularx}
\begin{tablenotes}
\item This table compares the performance of the three MII attacks: GS-MIIs, IS-MIIs, and RS-MIIs. Performance corresponds to the percentage of MIIs constructed that would be successfully verified with their corresponding $(\mathbf{p},\mathbf{q})_\text{pro}$ pair, based on an $\epsilon$ threshold. \emph{Italics} indicate the best performance per comparator.
\end{tablenotes}
\end{threeparttable}
\end{table}

IS-MIIs perform much better than GS-MIIs. Against the SENets less than half of IS-MII attacks succeed, whilst against LtNet$_{256}$ about two-thirds succeed. Note that since IS-MIIs do not make use of a comparator in their construction these differences in performance, depending on the comparator attacked, are not due to matched vs. similar vs. mismatched mode but reflect hidden differences in the vulnerability of the networks. It is noteworthy that these differences are not apparent from their performance on normal data which were similar (see \Cref{table:auroc_far}). 

RS-MIIs also perform much better than GS-MIIs. There is no clear winner comparing RS- and IS-MIIs. For matched and similar mode RS performs distinctly better than IS (77\% and 71\% vs. 37\% and 40\%), but for mismatched mode slightly worse (54\% vs. 66\%).

\section{Discussion}
\label{sec:discussion}
We now discuss issues arising especially with regards to improvement; first from the adversary's perspective, then from the defender's.

\subsection{Improved Multiple-Identity Image Attacks}
\subsubsection{Reducing Detectability}
GS-MIIs utilise real images that undergo no form of manipulation, and are therefore only detectable based on an MII being insufficiently similar to the appearance of the adversary and accomplice.

Contrastingly, there are clear and obvious cues of manipulation for IS- and RS-MIIs. The former suffer from visual ghosting artefacts; and the latter lack detail at the transition between the foreground (face region) and background. Clearly, more visually faultless MIIs can be constructed in both cases. For example, rather than generating MIIs, where the generator utilises the entire image, which includes the presence of identity irrelevant non-facial features such as the background, MIIs can instead be generated by binarising the image into facial and non-facial regions, such that only the facial area is manipulated prior to recombination with the background of one of the constituent images.

With specific regard to IS-MIIs, which rely on the precise specification of common image features for warping, advancements in automated landmark localisation (e.g. utilising a multi-task cascaded deep convolutional neural network~\cite{zhang2016joint}) are likely to offer improvements over the shallow landmark predictor used in this work, thus reducing ghosting. Nonetheless, generating photorealistic morphed images totally free from visual artefacts is still a challenging task. Heterogeneous, and intertwined, factors of variation stemming from pose, skin colour, hair style, etc. impact the realism of images generated on the continuum. That is, slight errors in landmark positioning inevitably give rise to ghosting, and these errors are common when a (shallow or deep) landmark predictor is learnt on a demographically biased dataset~\cite{mcduff2018identifying}. Therefore, combining a deep learning based landmark predictor (trained on an unbiased dataset) with both splice morphing~\cite{makrushin2017automatic} (considering only the facial region) and Poisson image editing~\cite{perez2003poisson} (smooth transition of low frequency details) appears to be the way forward. Note that regardless of improvements, IS-MIIs assume that the structural relationship between face images holds, i.e. that each face image is frontal, which is restrictive. Consequently, to generate IS-MIIs where this condition does not hold, an additional module will be required---e.g. a deep generative model capable of generating a novel (frontal) view of the face~\cite{huang2017beyond}. If, however, a deep generative neural network is going to employed, then it brings into question the utility of the entire IS-MII procedure, since one could instead employ RS-MIIs, which are capable of generating MIIs regardless of the pose of the reference images (when trained on face images where pose is a factor of variation).

With specific regard to RS-MIIs, an image $\mathbf{p}$ can be modelled as the composite of the foreground ($\mathbf{p}_F$) and background ($\mathbf{p}_B$) via a matting equation, i.e. ${\mathbf{p}=(1-\mathbf{G})\mathbf{p}_F + \mathbf{G}\mathbf{p}_B}$, where $\mathbf{G}$ is an occlusion matrix that determines the visibility of background pixels. For instance, \cite{yan2016attribute2image,shu2017neural} utilise a layered conditional generative model that generates images based on disentangled representations ${\widetilde{\mathbf{p}}=[\widetilde{\mathbf{p}}_{F},\widetilde{\mathbf{p}}_{B}]}$, where $\widetilde{\mathbf{p}}_{F}$ and $\widetilde{\mathbf{p}}_{B}$ capture the foreground and background factors of variation. This was shown to give visually pleasing results in the tasks of attribute-conditioned image reconstruction and completion. One negative, however, is that the method necessitates that the foreground layer $\mathbf{p}_F$ is observable during learning, i.e. the method in its current form does not work unsupervised. Alternatively, the foreground can be extracted utilising a \emph{copy-pasting} generative adversarial network recently proposed by \cite{arandjelovic2019object}, where objects are segmented (wholly unsupervised) for a given input image. The generator learns to discover an object in an image by compositing it into a different image, with the aim of fooling the discriminator into classifying the resulting image as real.

\subsubsection{Reducing the Gap to Ideal MIIs}
As it turned out, in \Cref{sec:constructing_miis_methods}, the MII construction methods evaluated produced suboptimal MIIs. This suggests that the studied attacks can be made stronger, so as to fulfil their theoretical potential. 

At present, the simplest solution for an adversary is to ensure that reference and live images are very similar, so as to increase the likelihood of an MII succeeding (assuming that the generator used is near ideal). However, this is clearly not always possible, especially as the time increases between when the reference images were captured and the live images.

With respect to GS-MIIs, although we do not show the dependency of their success rate on the gallery size---the larger the gallery is the better a match can be found. We have examined this and estimate that the gallery would need to be \num{\sim 1}B faces in order to achieve 50\% success. This renders GS-MIIs somewhat impractical, albeit far from impossible.

By considering the behaviour and performance of a hypothetical \emph{ideal} method for MII generation, an adversary could examine how well their method of MII generation realises this ideal. In particular, this can be done for IS- and RS-MII methods. Firstly, IS-MIIs can be made more ideal by optimising methods for landmarking, warping and interpolation with the ideal MII distance distribution in mind. Secondly, RS-MIIs can be made more ideal by perhaps employing a lower level intermediate representation that retains more image content, e.g. a convolutional layer, since the learnt filters in earlier layers are likely to be more similar across comparators than later layers which specialise to their task and training data.

It should be noted that at present we do not currently know to what extent the failures of photorealism are stochastically moving IS- and RS-MIIs away from their ideals. Speculatively, this \emph{may} have an effect, therefore improving MII photorealism would not only reduce their detectability, but could also assist in reducing the current gap between MIIs as they stand and their ideals.

\subsubsection{Choosing the Right Accomplice}
\begin{figure}[!t]
\centering
\includegraphics[width=1\columnwidth]{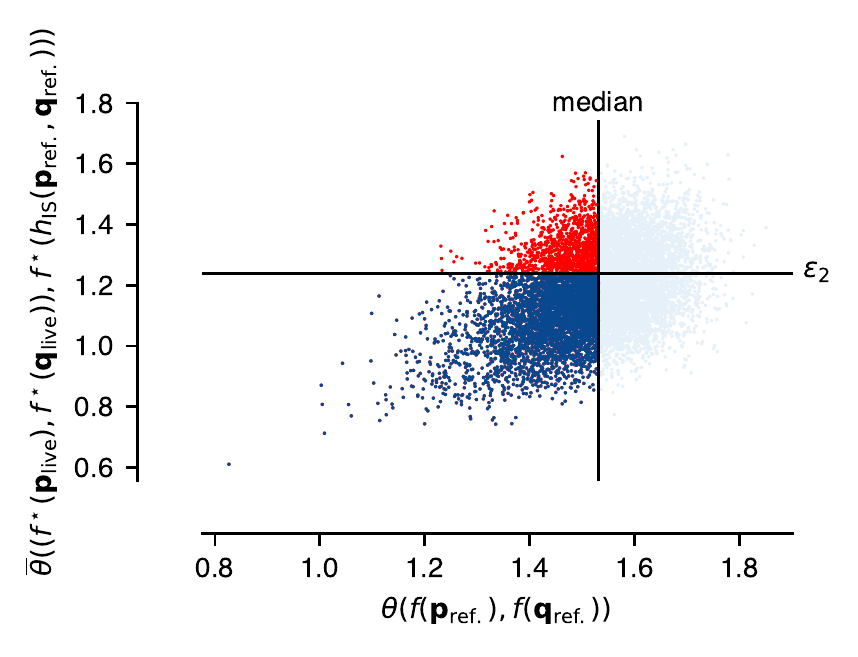}
\caption{Each point corresponds to a pair of identities $p$ and $q$. The horizontal axis is the angular distance between their reference images according to the adversary’s comparator SENet$_{256}$. The vertical axis is the MII distance according to the comparator being attacked, which here is LtNet$_{256}$ (mismatched with adversary). Points below the $\epsilon_2$ line are successful attacks against LtNet$_{256}$. Points to the left of the median line correspond to adversary-accomplice pairings that are fairly similar in appearance according to SENet$_{256}$. }  \label{fig:clever_attack}
\end{figure}

The ideal scenario for an adversary-accomplice pairing (e.g. $p$ and $q$) does not necessitate that they are initially visually similar in any regard. Throughout this paper, we represented this scenario by generating MIIs based on randomly sampled image pairs $(\mathbf{p}_{\text{ref}},\mathbf{q}_{\text{ref}})\in\mathcal{P}^{-}$. However, recall the scenario outlined in \Cref{sec:intro}, where individual $p$ wished to access some facility without their identity being detected. If $p$ works with a random collaborator $q$ to generate, for instance, an IS-MII, then they would have a ${\sim}66\%$ chance of success attacking LtNet$_{256}$. If, however, $p$ seeks out a collaborator $q$ who, to some degree, shares a resemblance, then the chance of success can be improved. For example, constraining $q$ to be within the $50\%$ of the population who are most visually similar to $p$ (LHS of the median in \Cref{fig:clever_attack}), then the success rate increases to ${\sim}77\%$ (lower left quadrant in \Cref{fig:clever_attack} as a proportion of the LHS of the median line). Clearly, further constraints would see further improvements.

\subsection{Improved Multiple-Identity Image Defences}
\subsubsection{Detecting MIIs}
Both our IS-MIIs and RS-MIIs have clear artefacts that a system could be trained to detect; but it is very doubtful whether such systems would be able to deal with improved versions of the MII generators. An \emph{arms race} is possible, but it is unlikely that regulated commercial systems would be able to keep up with the nimbleness of unregulated adversarial development. This is a consequence of the fact that most detection pipelines employ supervised machine learning algorithms, which have good generalisation ability when the training and testing datasets are sampled from the same distribution. Using deep learning methods, for example, super-human performance~\cite{he2015delving} can be achieved at recognising objects, places and people in consumer photography. This is possible because the statistics of this data are stable over time. In security applications, such as identity verification, the statistics of the rare set (MIIs) cannot be assumed to be stable, as adversaries seek to innovate and improve their methods. Bluntly, with sufficient examples of MIIs, as they exist in 2019, a supervised learning system that detected MIIs until 2020 might be possible, but it would likely fail in 2021. 

Anomaly (or novelty) detection systems, however, that utilise unsupervised learning, trained only on the normal dataset, would not have this inevitable obsolescence. Anomaly detection learns the range and dimensions of normal variation, so that suspicious deviations from this can be detected. For instance, a latent space autoregression framework~\cite{abati2019latent} based on \emph{memory} and \emph{surprisal} is viable, since an anomalous \emph{event} can be expressed in terms of one's capability to remember it (e.g.  sparse dictionary learning), as well as the degree to which the event is surprising (e.g. in terms of probability). We hypothesise that anomaly detection can be as effective at detecting known signs as a supervised approach, but will also be able to detect signs of improved MII attack methods.

\subsubsection{Securing the Comparator}
Our results in \Cref{sec:effectiveness} significantly highlight the following: (i) poisoning attacks optimised for one comparator can be very effective when transferred to a disparate comparator that was initially trained in the exact same way (prior to fine-tuning); and (ii) transferring attacks to completely different comparators (feature domains) with dissimilar architectures and training data \emph{can} be effective---contrary to the assertion made in \cite{scherhag2019face}. Furthermore, considering the attacks were randomly generated, LtNet$_{256}$ exhibits a substantial vulnerability to RS-MIIs optimised for SENet$_{128}$. Nevertheless, the results indicate that IS-MIIs are a more fruitful attack, at $\epsilon_2$, when one does not have access to a proxy face comparator that is \emph{similar} to the comparator that one wishes to attack---i.e. optimising RS-MIIs for SENet$_{128}$ so as to attack SENet$_{256}$.

The transferability of IS-MIIs and RS-MIIs should not come as a surprise, since the angular distance distributions $\mathcal{P}^{+}$ and $\mathcal{P}^{-}$, across the face comparators, exhibit a positive linear correlation (recall \Cref{table:correlation_coeffs}). Thus, to a degree, each comparator is roughly performing in a similar manner---as one would expect of discriminatively induced networks. Although the arrangement of the face representations may differ from comparator to comparator, if an MII is sufficiently similar in one feature domain to $\mathbf{p}_{\text{ref}}$ and $\mathbf{q}_{\text{ref}}$ then it is likely to also be similar in another. Therefore, securing a comparator is an insufficient defence to poisoning attacks.

\subsubsection{Improved Comparators}
Face comparators are known to be vulnerable to MII poisoning attacks~\cite{ferrara2014magic}, which is reinforced by our results. Research on this mode of attack has focused on defence~\cite{seibold2017detection,asaad2017topological,ferrara2018face,debiasi2018prnu,raghavendra2016detecting}, with little to no explanation given as to why face comparators are vulnerable.  Two aspects of our analysis suggest directions for improvement. 

First is the observation that some comparators are better than others against IS-MIIs (which do not make use of a comparator), even though they have similar performance on normal data. Understanding why may guide future development.

Second is what we observed with respect to the $\mathcal{P}^{+}$ and $\mathcal{P}^{-}$ distance distributions in \cref{fig:hists_cos_scores}, i.e. the lack of clear separation between the two. Comparators are typically learnt by minimising a supervised softmax-based loss~\cite{cao2018vggface2,wu2018light}, with abstract high-level intermediate representations extracted for the purposes of open-set verification. The discriminative ability of the representations is a by-product of the training process, and is not explicitly enforced, thus resulting in decision boundaries that do not directly maximise class separability beyond what is necessary for discrimination. However, the popularity of these approaches stems from the ease of their training, since, for example, more direct metric learning approaches~\cite{parkhi2015deep,schroff2015facenet}, which directly embed images into Euclidean space based on relative relationships amongst inputs (minimising intra-class variance and maximising inter-class variance) can be difficult to train: (i) constraints are typically exponentially large in number, but highly redundant, thus leading to slow convergence; and (ii) mining informative and non-redundant constraints to alleviate (i) is difficult and can lead to training instabilities.

To overcome some of the issues outlined above, improved softmax-based losses have been proposed. For example, those with the objective of: (i) penalising both the distance between a sample and its class centre, in representation space, as well as enforcing inter-class separability~\cite{wen2016discriminative}; (ii) penalising the distance between the centroid of each class~\cite{wu2017deep}; (iii) maximising the margin between the farthest intra-class sample and its nearest inter-class neighbour~\cite{deng2017marginal}; or (iv) employing angular-based margins~\cite{liu2017sphereface,wang2018additive,wang2018cosface}, e.g.  directly optimising the geodesic distance margin on the hypersphere, so as to enforce a larger margin between classes~\cite{deng2018arcface}. Nonetheless, our distributional hypothesis still stands for such learning paradigms. The primary difference being that direct methods may lead to lower $\epsilon$ verification thresholds, since the distributions of angular distances of pairs in $\mathcal{P}^{+}$ and $\mathcal{P}^{-}$ is expected to have greater separation, thus potentially leading to the rejection of a greater number of MIIs. However, since inter-class separability is explicitly enforced, it becomes increasingly likely that two random face images of different persons will be orthogonal, which by implication means their optimal MII is almost surely at an angular distance of $\pi/4$. Therefore, as long as the $\epsilon$ threshold employed lies to the right of the distribution of angular distances attained using idealised MIIs (due to insufficiently reducing the intra-class variance) with mean $\pi/4$, then a comparator remains theoretically vulnerable to randomly constructed poisoning attacks. 

Patently, the key to mitigating against randomised poisoning attacks lies in reducing intra-class variance, especially in light of the fact that algorithms for constructing MII attacks will most likely continue to improve (reaching their theorised potential). One solution would be to indefinitely contract samples from the same class during comparator training~\cite{simo2015discriminative,chopra2005learning}---by way of \emph{contrastive} pairwise metric learning. That is, matching pairs continue to contribute to the loss, in an attempt to drive their angular distances to zero---beyond the point needed to differentiate them from dissimilarly labelled samples. However, this method of learning is known to cause overfitting. Consequently, a balance must be struck between robustness to MIIs and generalisation ability for open-set verification. An adapted solution would be to ensure that the $\mathcal{P}^{+}$ distribution of distances is well separated from the ideal MII distribution of distances, as opposed to the $\mathcal{P}^{-}$ distribution of distances, employing for instance a histogram loss~\cite{ustinova2016learning}.

\section{Conclusion}
\label{sec:conclusion}
Face comparators are known to be vulnerable to MII poisoning attacks, which is reinforced by our results. However, research on this mode of attack has focused on defence, with little to no explanation given as to why face comparators are vulnerable. In contrast, we provided an intuitive view on the role of the face representation spaces used for verification, arguing that the principal cause of the vulnerability is that representations of different identity faces are always close to orthogonal, and the representations of same identity faces are only modestly closer than orthogonal. This is sufficient for open-set verification on normal data but provides an opportunity for MII attacks. Importantly, by considering the behaviour and performance of a hypothetical \emph{ideal} method for MII generation, we were able to examine how well existing MII generators realise theoretically optimal MIIs---permitting one to establish the vulnerability of a specific comparator to MIIs. In addition, we showed that transference of MIIs from one comparator to another is made possible due to the representation spaces of dissimilar comparators being sufficiently similar, as such securing a comparator is an insufficient defence.

It is unclear whether MIIs can be completely defended against---although it is unlikely that any of our generated MIIs (in \Cref{fig:methods_comp}) would be mistaken for the constituent identities by a human expert, we are not confident that this would still be true for improved generation systems that get closer to the ideal.


\ifCLASSOPTIONcaptionsoff
  \newpage
\fi



\bibliographystyle{IEEEtran}

\begin{thebibliography}{10}
\providecommand{\url}[1]{#1}
\csname url@samestyle\endcsname
\providecommand{\newblock}{\relax}
\providecommand{\bibinfo}[2]{#2}
\providecommand{\BIBentrySTDinterwordspacing}{\spaceskip=0pt\relax}
\providecommand{\BIBentryALTinterwordstretchfactor}{4}
\providecommand{\BIBentryALTinterwordspacing}{\spaceskip=\fontdimen2\font plus
\BIBentryALTinterwordstretchfactor\fontdimen3\font minus
  \fontdimen4\font\relax}
\providecommand{\BIBforeignlanguage}[2]{{%
\expandafter\ifx\csname l@#1\endcsname\relax
\typeout{** WARNING: IEEEtran.bst: No hyphenation pattern has been}%
\typeout{** loaded for the language `#1'. Using the pattern for}%
\typeout{** the default language instead.}%
\else
\language=\csname l@#1\endcsname
\fi
#2}}
\providecommand{\BIBdecl}{\relax}
\BIBdecl

\bibitem{schroff2015facenet}
F.~Schroff, D.~Kalenichenko, and J.~Philbin, ``Facenet: A unified embedding for
  face recognition and clustering,'' in \emph{Proceedings of the IEEE
  conference on computer vision and pattern recognition}, 2015, pp. 815--823.

\bibitem{parkhi2015deep}
O.~M. Parkhi, A.~Vedaldi, A.~Zisserman \emph{et~al.}, ``Deep face
  recognition.'' in \emph{bmvc}, vol.~1, no.~3, 2015, p.~6.

\bibitem{liu2017sphereface}
W.~Liu, Y.~Wen, Z.~Yu, M.~Li, B.~Raj, and L.~Song, ``Sphereface: Deep
  hypersphere embedding for face recognition,'' in \emph{The IEEE Conference on
  Computer Vision and Pattern Recognition (CVPR)}, vol.~1, 2017, p.~1.

\bibitem{cao2018vggface2}
Q.~Cao, L.~Shen, W.~Xie, O.~M. Parkhi, and A.~Zisserman, ``Vggface2: A dataset
  for recognising faces across pose and age,'' in \emph{2018 13th IEEE
  International Conference on Automatic Face \& Gesture Recognition (FG
  2018)}.\hskip 1em plus 0.5em minus 0.4em\relax IEEE, 2018, pp. 67--74.

\bibitem{wu2018light}
X.~Wu, R.~He, Z.~Sun, and T.~Tan, ``A light cnn for deep face representation
  with noisy labels,'' \emph{IEEE Transactions on Information Forensics and
  Security}, vol.~13, no.~11, pp. 2884--2896, 2018.

\bibitem{jain201650}
A.~K. Jain, K.~Nandakumar, and A.~Ross, ``50 years of biometric research:
  Accomplishments, challenges, and opportunities,'' \emph{Pattern Recognition
  Letters}, vol.~79, pp. 80--105, 2016.

\bibitem{ratha2001enhancing}
N.~K. Ratha, J.~H. Connell, and R.~M. Bolle, ``Enhancing security and privacy
  in biometrics-based authentication systems,'' \emph{IBM systems Journal},
  vol.~40, no.~3, pp. 614--634, 2001.

\bibitem{wolberg1998image}
G.~Wolberg, ``Image morphing: a survey,'' \emph{The visual computer}, vol.~14,
  no.~8, pp. 360--372, 1998.

\bibitem{ferrara2014magic}
M.~Ferrara, A.~Franco, and D.~Maltoni, ``The magic passport,'' in
  \emph{Biometrics (IJCB), 2014 IEEE International Joint Conference on}.\hskip
  1em plus 0.5em minus 0.4em\relax IEEE, 2014, pp. 1--7.

\bibitem{scherhag2017vulnerability}
U.~Scherhag, R.~Raghavendra, K.~B. Raja, M.~Gomez-Barrero, C.~Rathgeb, and
  C.~Busch, ``On the vulnerability of face recognition systems towards morphed
  face attacks,'' in \emph{2017 5th International Workshop on Biometrics and
  Forensics (IWBF)}.\hskip 1em plus 0.5em minus 0.4em\relax IEEE, 2017, pp.
  1--6.

\bibitem{robertson2018detecting}
D.~J. Robertson, A.~Mungall, D.~G. Watson, K.~A. Wade, S.~J. Nightingale, and
  S.~Butler, ``Detecting morphed passport photos: a training and individual
  differences approach,'' \emph{Cognitive research: principles and
  implications}, vol.~3, no.~1, p.~27, 2018.

\bibitem{seibold2018reflection}
C.~Seibold, A.~Hilsmann, and P.~Eisert, ``Reflection analysis for face morphing
  attack detection,'' \emph{arXiv preprint arXiv:1807.02030}, 2018.

\bibitem{makrushin2017automatic}
A.~Makrushin, T.~Neubert, and J.~Dittmann, ``Automatic generation and detection
  of visually faultless facial morphs.'' in \emph{VISIGRAPP (6: VISAPP)}, 2017,
  pp. 39--50.

\bibitem{vyas2015automatic}
J.~P. Vyas, M.~V. Joshi, and M.~S. Raval, ``Automatic target image detection
  for morphing,'' \emph{Journal of Visual Communication and Image
  Representation}, vol.~27, pp. 28--43, 2015.

\bibitem{perez2003poisson}
P.~P{\'e}rez, M.~Gangnet, and A.~Blake, ``Poisson image editing,'' \emph{ACM
  Transactions on graphics (TOG)}, vol.~22, no.~3, pp. 313--318, 2003.

\bibitem{debiasi2018prnu}
L.~Debiasi, U.~Scherhag, C.~Rathgeb, A.~Uhl, and C.~Busch, ``Prnu-based
  detection of morphed face images,'' in \emph{2018 International Workshop on
  Biometrics and Forensics (IWBF)}.\hskip 1em plus 0.5em minus 0.4em\relax
  IEEE, 2018, pp. 1--7.

\bibitem{mahendran2015understanding}
A.~Mahendran and A.~Vedaldi, ``Understanding deep image representations by
  inverting them,'' in \emph{Proceedings of the IEEE conference on computer
  vision and pattern recognition}, 2015, pp. 5188--5196.

\bibitem{dosovitskiy2016inverting}
A.~Dosovitskiy and T.~Brox, ``Inverting visual representations with
  convolutional networks,'' in \emph{Proceedings of the IEEE Conference on
  Computer Vision and Pattern Recognition}, 2016, pp. 4829--4837.

\bibitem{dosovitskiy2016generating}
------, ``Generating images with perceptual similarity metrics based on deep
  networks,'' in \emph{Advances in neural information processing systems},
  2016, pp. 658--666.

\bibitem{shu2017neural}
Z.~Shu, E.~Yumer, S.~Hadap, K.~Sunkavalli, E.~Shechtman, and D.~Samaras,
  ``Neural face editing with intrinsic image disentangling,'' in
  \emph{Proceedings of the IEEE Conference on Computer Vision and Pattern
  Recognition}, 2017, pp. 5541--5550.

\bibitem{perarnau2016invertible}
G.~Perarnau, J.~Van De~Weijer, B.~Raducanu, and J.~M. {\'A}lvarez, ``Invertible
  conditional gans for image editing,'' \emph{arXiv preprint arXiv:1611.06355},
  2016.

\bibitem{lample2017fader}
G.~Lample, N.~Zeghidour, N.~Usunier, A.~Bordes, L.~Denoyer \emph{et~al.},
  ``Fader networks: Manipulating images by sliding attributes,'' in
  \emph{Advances in Neural Information Processing Systems}, 2017, pp.
  5967--5976.

\bibitem{he2017arbitrary}
Z.~He, W.~Zuo, M.~Kan, S.~Shan, and X.~Chen, ``Arbitrary facial attribute
  editing: Only change what you want,'' \emph{arXiv preprint arXiv:1711.10678},
  vol.~1, no.~3, 2017.

\bibitem{wang2018face}
Z.~Wang, X.~Tang, W.~Luo, and S.~Gao, ``Face aging with identity-preserved
  conditional generative adversarial networks,'' in \emph{Proceedings of the
  IEEE Conference on Computer Vision and Pattern Recognition}, 2018, pp.
  7939--7947.

\bibitem{damer2019morgan}
N.~Damer, A.~M. Saladi{\'e}, A.~Braun, and A.~Kuijper, ``Morgan: Recognition
  vulnerability and attack detectability of face morphing attacks created by
  generative adversarial network,'' in \emph{2018 IEEE 9th International
  Conference on Biometrics Theory, Applications and Systems (BTAS)}.\hskip 1em
  plus 0.5em minus 0.4em\relax IEEE, 2019, pp. 1--10.

\bibitem{scherhag2019face}
U.~Scherhag, C.~Rathgeb, J.~Merkle, R.~Breithaupt, and C.~Busch, ``Face
  recognition systems under morphing attacks: A survey,'' \emph{IEEE Access},
  vol.~7, pp. 23\,012--23\,026, 2019.

\bibitem{seibold2017detection}
C.~Seibold, W.~Samek, A.~Hilsmann, and P.~Eisert, ``Detection of face morphing
  attacks by deep learning,'' in \emph{International Workshop on Digital
  Watermarking}.\hskip 1em plus 0.5em minus 0.4em\relax Springer, 2017, pp.
  107--120.

\bibitem{asaad2017topological}
A.~Asaad and S.~Jassim, ``Topological data analysis for image tampering
  detection,'' in \emph{International Workshop on Digital Watermarking}.\hskip
  1em plus 0.5em minus 0.4em\relax Springer, 2017, pp. 136--146.

\bibitem{ferrara2018face}
M.~Ferrara, A.~Franco, and D.~Maltoni, ``Face demorphing,'' \emph{IEEE
  Transactions on Information Forensics and Security}, vol.~13, no.~4, pp.
  1008--1017, 2018.

\bibitem{raghavendra2016detecting}
R.~Raghavendra, K.~B. Raja, and C.~Busch, ``Detecting morphed face images,'' in
  \emph{Biometrics Theory, Applications and Systems (BTAS), 2016 IEEE 8th
  International Conference on}.\hskip 1em plus 0.5em minus 0.4em\relax IEEE,
  2016, pp. 1--7.

\bibitem{birajdar2013digital}
G.~K. Birajdar and V.~H. Mankar, ``Digital image forgery detection using
  passive techniques: A survey,'' \emph{Digital Investigation}, vol.~10, no.~3,
  pp. 226--245, 2013.

\bibitem{scherhag2018performance}
U.~Scherhag, C.~Rathgeb, and C.~Busch, ``Performance variation of morphed face
  image detection algorithms across different datasets,'' in \emph{2018
  International Workshop on Biometrics and Forensics (IWBF)}.\hskip 1em plus
  0.5em minus 0.4em\relax IEEE, 2018, pp. 1--6.

\bibitem{spreeuwers2018towards}
L.~Spreeuwers, M.~Schils, and R.~Veldhuis, ``Towards robust evaluation of face
  morphing detection,'' in \emph{2018 26th European Signal Processing
  Conference (EUSIPCO)}.\hskip 1em plus 0.5em minus 0.4em\relax IEEE, 2018, pp.
  1027--1031.

\bibitem{scherhag2018towards}
U.~Scherhag, C.~Rathgeb, and C.~Busch, ``Towards detection of morphed face
  images in electronic travel documents,'' in \emph{2018 13th IAPR
  International Workshop on Document Analysis Systems (DAS)}.\hskip 1em plus
  0.5em minus 0.4em\relax IEEE, 2018, pp. 187--192.

\bibitem{kraetzer2017modeling}
C.~Kraetzer, A.~Makrushin, T.~Neubert, M.~Hildebrandt, and J.~Dittmann,
  ``Modeling attacks on photo-id documents and applying media forensics for the
  detection of facial morphing,'' in \emph{Proceedings of the 5th ACM Workshop
  on Information Hiding and Multimedia Security}.\hskip 1em plus 0.5em minus
  0.4em\relax ACM, 2017, pp. 21--32.

\bibitem{neubert2018extended}
T.~Neubert, A.~Makrushin, M.~Hildebrandt, C.~Kraetzer, and J.~Dittmann,
  ``Extended stirtrace benchmarking of biometric and forensic qualities of
  morphed face images,'' \emph{IET Biometrics}, vol.~7, no.~4, pp. 325--332,
  2018.

\bibitem{scherhag2018detecting}
U.~Scherhag, D.~Budhrani, M.~Gomez-Barrero, and C.~Busch, ``Detecting morphed
  face images using facial landmarks,'' in \emph{International Conference on
  Image and Signal Processing}.\hskip 1em plus 0.5em minus 0.4em\relax
  Springer, 2018, pp. 444--452.

\bibitem{phillips1997feret}
P.~J. Phillips, H.~Moon, P.~Rauss, and S.~A. Rizvi, ``The feret evaluation
  methodology for face-recognition algorithms,'' in \emph{Proceedings of IEEE
  Computer Society Conference on Computer Vision and Pattern
  Recognition}.\hskip 1em plus 0.5em minus 0.4em\relax IEEE, 1997, pp.
  137--143.

\bibitem{karras2018style}
T.~Karras, S.~Laine, and T.~Aila, ``A style-based generator architecture for
  generative adversarial networks,'' \emph{arXiv preprint arXiv:1812.04948},
  2018.

\bibitem{king2009dlib}
D.~E. King, ``Dlib-ml: A machine learning toolkit,'' \emph{Journal of Machine
  Learning Research}, vol.~10, no. Jul, pp. 1755--1758, 2009.

\bibitem{kazemi2014one}
V.~Kazemi and J.~Sullivan, ``One millisecond face alignment with an ensemble of
  regression trees,'' in \emph{Proceedings of the IEEE conference on computer
  vision and pattern recognition}, 2014, pp. 1867--1874.

\bibitem{hu2018squeeze}
J.~Hu, L.~Shen, and G.~Sun, ``Squeeze-and-excitation networks,'' in
  \emph{Proceedings of the IEEE conference on computer vision and pattern
  recognition}, 2018, pp. 7132--7141.

\bibitem{he2016deep}
K.~He, X.~Zhang, S.~Ren, and J.~Sun, ``Deep residual learning for image
  recognition,'' in \emph{Proceedings of the IEEE conference on computer vision
  and pattern recognition}, 2016, pp. 770--778.

\bibitem{guo2016ms}
Y.~Guo, L.~Zhang, Y.~Hu, X.~He, and J.~Gao, ``Ms-celeb-1m: A dataset and
  benchmark for large-scale face recognition,'' in \emph{European Conference on
  Computer Vision}.\hskip 1em plus 0.5em minus 0.4em\relax Springer, 2016, pp.
  87--102.

\bibitem{nair2010rectified}
V.~Nair and G.~E. Hinton, ``Rectified linear units improve restricted boltzmann
  machines,'' in \emph{Proceedings of the 27th international conference on
  machine learning (ICML-10)}, 2010, pp. 807--814.

\bibitem{goodfellow2013maxout}
I.~J. Goodfellow, D.~Warde-Farley, M.~Mirza, A.~Courville, and Y.~Bengio,
  ``Maxout networks,'' \emph{arXiv preprint arXiv:1302.4389}, 2013.

\bibitem{yi2014learning}
D.~Yi, Z.~Lei, S.~Liao, and S.~Z. Li, ``Learning face representation from
  scratch,'' \emph{arXiv preprint arXiv:1411.7923}, 2014.

\bibitem{delaunay1934sphere}
B.~Delaunay, ``Sur la sph{\`e}re vide,'' \emph{Bulletin of Academy of Sciences
  of the USSR}, pp. 793--800, 1934.

\bibitem{isola2017image}
P.~Isola, J.-Y. Zhu, T.~Zhou, and A.~A. Efros, ``Image-to-image translation
  with conditional adversarial networks,'' in \emph{Proceedings of the IEEE
  conference on computer vision and pattern recognition}, 2017, pp. 1125--1134.

\bibitem{mao2017least}
X.~Mao, Q.~Li, H.~Xie, R.~Y. Lau, Z.~Wang, and S.~Paul~Smolley, ``Least squares
  generative adversarial networks,'' in \emph{Proceedings of the IEEE
  International Conference on Computer Vision}, 2017, pp. 2794--2802.

\bibitem{zhang2016joint}
K.~Zhang, Z.~Zhang, Z.~Li, and Y.~Qiao, ``Joint face detection and alignment
  using multitask cascaded convolutional networks,'' \emph{IEEE Signal
  Processing Letters}, vol.~23, no.~10, pp. 1499--1503, 2016.

\bibitem{mcduff2018identifying}
D.~McDuff, R.~Cheng, and A.~Kapoor, ``Identifying bias in ai using
  simulation,'' \emph{arXiv preprint arXiv:1810.00471}, 2018.

\bibitem{huang2017beyond}
R.~Huang, S.~Zhang, T.~Li, and R.~He, ``Beyond face rotation: Global and local
  perception gan for photorealistic and identity preserving frontal view
  synthesis,'' in \emph{Proceedings of the IEEE International Conference on
  Computer Vision}, 2017, pp. 2439--2448.

\bibitem{yan2016attribute2image}
X.~Yan, J.~Yang, K.~Sohn, and H.~Lee, ``Attribute2image: Conditional image
  generation from visual attributes,'' in \emph{European Conference on Computer
  Vision}.\hskip 1em plus 0.5em minus 0.4em\relax Springer, 2016, pp. 776--791.

\bibitem{arandjelovic2019object}
R.~Arandjelovi{\'c} and A.~Zisserman, ``Object discovery with a copy-pasting
  gan,'' \emph{arXiv preprint arXiv:1905.11369}, 2019.

\bibitem{he2015delving}
K.~He, X.~Zhang, S.~Ren, and J.~Sun, ``Delving deep into rectifiers: Surpassing
  human-level performance on imagenet classification,'' in \emph{Proceedings of
  the IEEE international conference on computer vision}, 2015, pp. 1026--1034.

\bibitem{abati2019latent}
D.~Abati, A.~Porrello, S.~Calderara, and R.~Cucchiara, ``Latent space
  autoregression for novelty detection,'' in \emph{International Conference on
  Computer Vision and Pattern Recognition}, 2019.

\bibitem{wen2016discriminative}
Y.~Wen, K.~Zhang, Z.~Li, and Y.~Qiao, ``A discriminative feature learning
  approach for deep face recognition,'' in \emph{European conference on
  computer vision}.\hskip 1em plus 0.5em minus 0.4em\relax Springer, 2016, pp.
  499--515.

\bibitem{wu2017deep}
Y.~Wu, H.~Liu, J.~Li, and Y.~Fu, ``Deep face recognition with center invariant
  loss,'' in \emph{Proceedings of the on Thematic Workshops of ACM Multimedia
  2017}.\hskip 1em plus 0.5em minus 0.4em\relax ACM, 2017, pp. 408--414.

\bibitem{deng2017marginal}
J.~Deng, Y.~Zhou, and S.~Zafeiriou, ``Marginal loss for deep face
  recognition,'' in \emph{Proceedings of the IEEE Conference on Computer Vision
  and Pattern Recognition Workshops}, 2017, pp. 60--68.

\bibitem{wang2018additive}
F.~Wang, J.~Cheng, W.~Liu, and H.~Liu, ``Additive margin softmax for face
  verification,'' \emph{IEEE Signal Processing Letters}, vol.~25, no.~7, pp.
  926--930, 2018.

\bibitem{wang2018cosface}
H.~Wang, Y.~Wang, Z.~Zhou, X.~Ji, D.~Gong, J.~Zhou, Z.~Li, and W.~Liu,
  ``Cosface: Large margin cosine loss for deep face recognition,'' in
  \emph{Proceedings of the IEEE Conference on Computer Vision and Pattern
  Recognition}, 2018, pp. 5265--5274.

\bibitem{deng2018arcface}
J.~Deng, J.~Guo, N.~Xue, and S.~Zafeiriou, ``Arcface: Additive angular margin
  loss for deep face recognition,'' \emph{arXiv preprint arXiv:1801.07698},
  2018.

\bibitem{simo2015discriminative}
E.~Simo-Serra, E.~Trulls, L.~Ferraz, I.~Kokkinos, P.~Fua, and F.~Moreno-Noguer,
  ``Discriminative learning of deep convolutional feature point descriptors,''
  in \emph{Proceedings of the IEEE International Conference on Computer
  Vision}, 2015, pp. 118--126.

\bibitem{chopra2005learning}
S.~Chopra, R.~Hadsell, Y.~LeCun \emph{et~al.}, ``Learning a similarity metric
  discriminatively, with application to face verification,'' in \emph{CVPR
  (1)}, 2005, pp. 539--546.

\bibitem{ustinova2016learning}
E.~Ustinova and V.~Lempitsky, ``Learning deep embeddings with histogram loss,''
  in \emph{Advances in Neural Information Processing Systems}, 2016, pp.
  4170--4178.

\end{thebibliography}
\end{document}